\documentclass[runningheads]{llncs}

\usepackage{eccv}

\usepackage{eccvabbrv}

\usepackage{graphicx}
\usepackage{booktabs}

\usepackage[accsupp]{axessibility}  

\usepackage{hyperref}

\usepackage{orcidlink}
\usepackage{enumitem}
\usepackage{bm}
\usepackage{multirow}
\newcommand{\modelname}[0]{NeuroPictor }

\begin{document}

\title{ NeuroPictor: Refining fMRI-to-Image Reconstruction via  Multi-individual Pretraining and Multi-level Modulation}

\titlerunning{NeuroPictor}

\author{Jingyang Huo$^{\star}$ \and Yikai Wang$^{\star}$ \and Yun Wang$^{\star}$ \and Xuelin Qian \and Chong Li  \and \\  Yanwei Fu$^{\dagger}$ \and Jianfeng Feng }

\authorrunning{J. Huo et al.}

\institute{Fudan University, Shanghai, China \\
\email{jyhuo22@m.fudan.edu.cn,  yanweifu@fudan.edu.cn}}

{\let\thefootnote\relax\footnotetext{
$^{\star}$ Equal contributions. \\
$^{\dagger}$ Corresponding author.}}

\maketitle

\begin{center}
\includegraphics[width=1.0\linewidth]{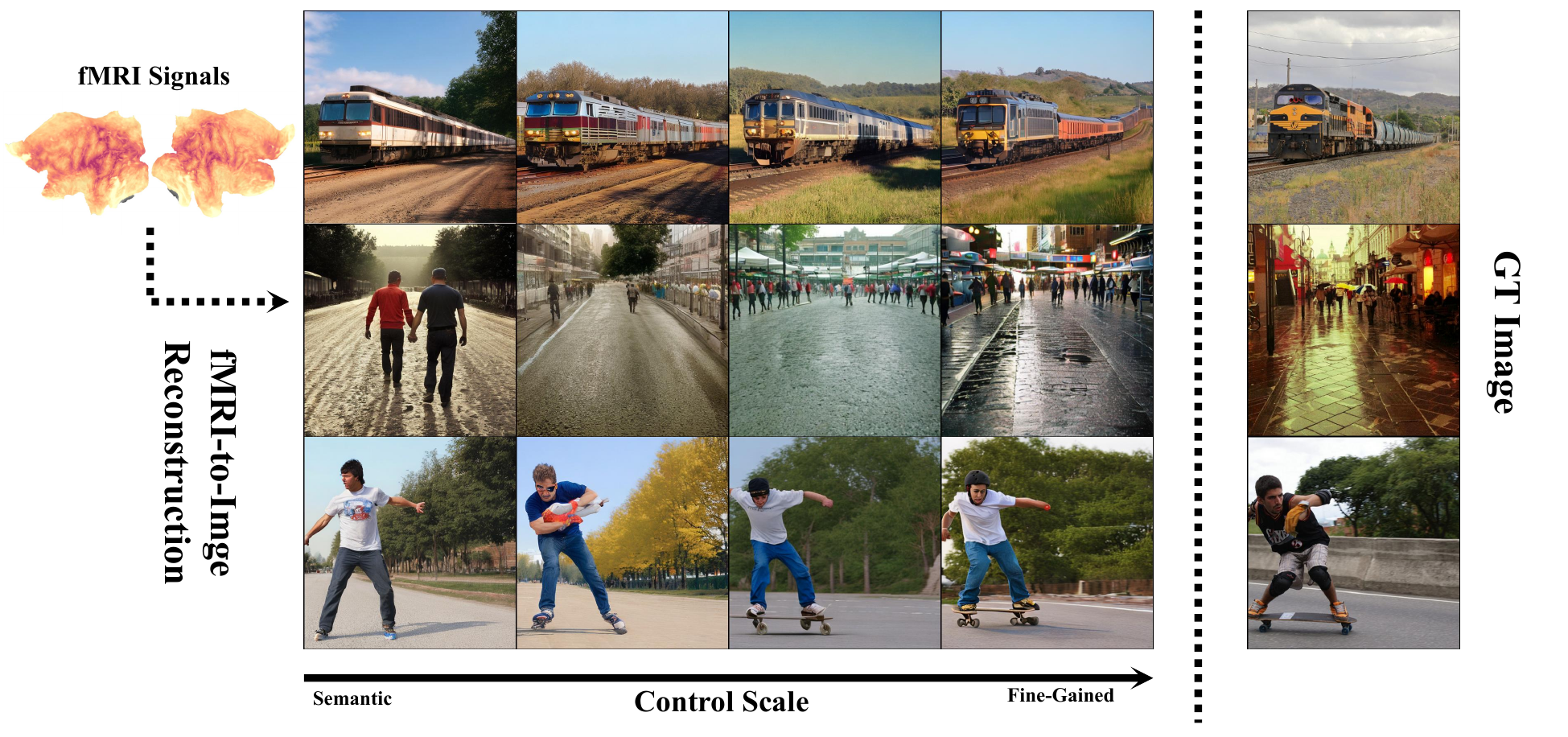}
\captionof{figure}{
\small 
NeuroPictor achieves precise control over decoding low-level structures from fMRI signals while preserving high-level semantics. The decoded images progress from reconstructing visual stimulus solely from high-level semantics to both high-level semantics and low-level structures as the influence increases from left to right. 
 \label{fig:teaser}}
\end{center}

\begin{abstract}
Recent fMRI-to-image approaches mainly focused on associating fMRI signals with specific conditions of pre-trained diffusion models.
These approaches, while producing high-quality images, capture only a limited aspect of the complex information in fMRI signals and offer little detailed control over image creation.
In contrast, this paper proposes to directly modulate the generation process of diffusion models using fMRI signals. 
Our approach, NeuroPictor, divides the fMRI-to-image process into three steps: 
i) fMRI calibrated-encoding, to tackle multi-individual pre-training for a shared latent space to minimize individual difference and enable the subsequent multi-subject training; 
ii) fMRI-to-image multi-subject pre-training, perceptually learning to guide diffusion model with high- and low-level conditions across different individuals;
iii) fMRI-to-image single-subject refining, similar with step ii but focus on adapting to particular individual. 
NeuroPictor extracts high-level semantic features from fMRI signals that characterizing the visual stimulus and incrementally fine-tunes the diffusion model with a low-level manipulation network to provide precise structural instructions. 
By training with about 67,000 fMRI-image pairs from various individuals, our model enjoys superior fMRI-to-image decoding capacity, particularly in the within-subject setting, as evidenced in benchmark datasets. Our code and model are available at \url{https://jingyanghuo.github.io/neuropictor/}.
  \keywords{Neural Decoding \and FMRI-to-Image \and Diffusion Model}
\end{abstract}

\section{Introduction}
\label{sec:intro}

Decoding visual stimuli perceived by the eyes and recorded by our brains is a fascinating task, serving as a bridge between two closely linked yet distinct fields: computer vision and visual neuroscience. In neuroscience, we rely on functional magnetic resonance imaging (fMRI), a non-invasive neuroimaging technique that detects brain activity by measuring changes in blood flow, offering valuable insights into cognitive processes and brain function~\cite{glover2011overview}. Meanwhile, the generative power of the Diffusion model enables the creation of visually appealing images from prior conditions or random signals, making it intriguing to decode fMRI-based brain activities and reconstruct images, as explored in~\cite{seung2000manifold}.

This paper explores the process of decoding fMRI signals into images. We outline the experimental pipeline in three stages: "See it. Say it. Sorted". Essentially, an individual unfamiliar with our task is presented with an image ("See it"). They then take a moment to reflect on this image ("Say it"), during which their brain activity is scanned and recorded using fMRI equipment. Finally, our algorithm uses the recorded fMRI signals to reconstruct the original image ("Sorted"). 
Most previous works~\cite{chen2022seeing,qian2023fmri}, focus on individual-specific fMRI-to-image decoders. Unfortunately, 
due to the potential unique brain activation patterns each brain processes information during the "say it" phase, there is no direct empirical evidence to show that another individual's fMRI data would assist in decoding for a specific individual.

Furthermore, in the "sorted" stage, researchers often employ latent codes extracted from fMRI signals to guide image generation through the Stable Diffusion (SD) model. 
Nevertheless, these codes tend to prioritize high-level semantics over low-level details, as fMRI voxels are aligned with CLIP text~\cite{ozcelik2023natural} and image features~\cite{scotti2023reconstructing} to capture semantic information.  
However, using latent codes directly for SD presents challenges due to the fundamental mismatch between the separately trained fMRI encoder/decoder for 1D signals and the SD model for 2D images.
Consequently, imperfections in both the fMRI signal-to-latent and latent-to-image projections lead to inevitable information loss and error accumulation, resulting in blurry reconstructions with lacking clear structure and pose information.

\begin{figure}[t] \small
\begin{centering}
\includegraphics[width=0.9\linewidth]{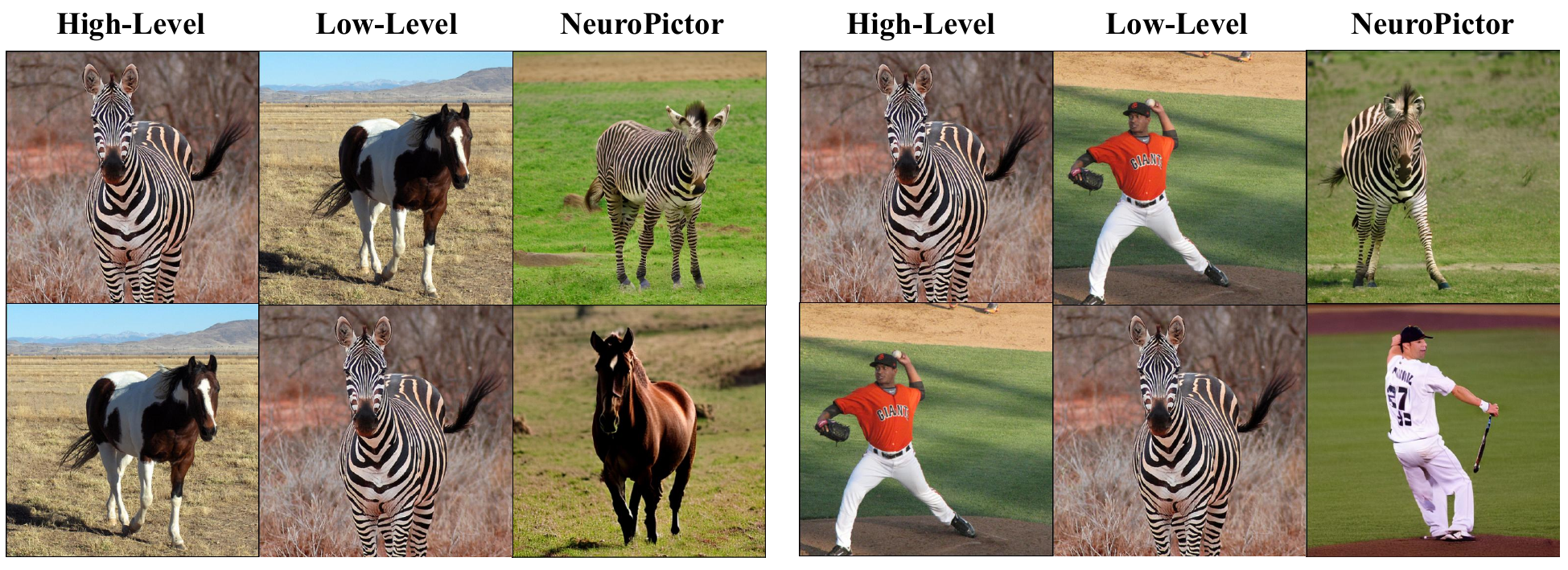}
\par\end{centering}
 \caption{ \small NeuroPictor can swap high-level fMRI features to manipulate image semantics while maintaining structural consistency.
\label{fig:switch}}
\end{figure}

To address these challenges, we introduce NeuroPictor\footnote{ {\small The name combines "Neuro" (representing human neurons) with "Pictor" (an easel for painting), symbolizing the depiction of neuron status.}}, a novel framework designed to refine fMRI-to-image reconstruction through multi-individual pretraining and multi-level modulation. Unlike previous approaches~\cite{chen2022seeing,qian2023fmri}  focusing on individual-specific decoders, NeuroPictor significantly improves fMRI-to-image decoding for single individuals by leveraging pretraining on multiple-individual fMRI-image pairs. 
Our method establishes a universal fMRI latent space to capture diverse neural signal information across subjects. Within this framework, we incorporate both low-level manipulation and high-level guiding networks as in Fig.~\ref{fig:teaser}.
The low-level manipulation network precisely adjusts the diffusion model's features, integrating complex brain signal information and eliminating the need for partial representation with intermediate images. 
Meanwhile, the high-level guiding network rectifies semantic gaps between text captions and visual images. 
These components form a unified pipeline, enabling training the full model from fMRI signals to visual stimulus directly via the fMRI-to-image reconstruction supervision and preventing information degradation and error accumulation from isolated processing.
Moreover, disentangling high-level meaning and low-level structure in fMRI signals opens up intriguing possibilities. For example, we can swap the high-level features of one fMRI signal with another, enabling precise semantic manipulation on the reconstructed image while preserving the structures, as depicted in Fig.~\ref{fig:switch}.

Formally, our NeuroPictor initially learns a universal fMRI latent space across individuals through multi-individual pre-training to capture all original signal information and individual differences. We use an fMRI encoder-decoder architecture to reconstruct the fMRI signal as 2D brain activation flatmaps. Afterward, we utilize the encoder to map the fMRI signal into latent codes. We introduce a High-Level Guiding Network (HLGN) to extract semantic-related signals essential in fMRI representations. The HLGN includes dual encoders aligning with text information and providing supplementary data to enhance generative model adaptability. Additionally, we incorporate a Low-Level Manipulation Network (LLMN) to collaborate with the diffusion model by directly manipulating underlying features. Recent studies \cite{zhang2023adding, mou2023t2i} have shown advancements through feature map adjustments in generative models to control fine-grained visual details. Our LLMN combines a feature transformation technique with a U-net Encoder to guide image generation by updating feature maps. By decoupling the decoding process and separately controlling high-level semantics and low-level manipulation, our NeuroPictor achieves precise fMRI-to-image reconstruction.

We propose to learn fMRI-to-image decoding in three steps.
i) The fMRI calibrated-encoding step.
We pre-train the fMRI signal processing model for fMRI reconstruction, calibrating fMRI signals of different individuals to a shared latent space.
ii) The fMRI-to-image multi-subject pre-training step.
We pre-train our full model directly for fMRI-to-image decoding, leveraging about 67k fMRI-image pairs from different individuals.
iii) For particular individual, we perform single-subject refining, using the same training strategy of step ii but focues on the refinement on the single subject.
Our results demonstrate the benefits of multi-individual pre-training for within-subject image reconstruction and the potential of the foundation fMRI decoding model. 
Extensive experiments validate the efficacy of our NeuroPictor.

\noindent\textbf{Contribution}. Our contributions are listed as follows:
(1) We propose pre-training on multiple-individual fMRI-image pairs to enhence fMRI-to-image decoding for single individuals;
(2) We capture diverse neural signal information and divide it into high-level and low-level guidance to supervise the diffusion generation process;
(3) We integrate all the components as a unified model for multi-subject pre-training and single-individual refining, adapting the full model directly based on the supervision of fMRI-to-image decoding.

\section{Related Work}
\label{sec:related}

\noindent
{\bf fMRI-to-image Reconstruction.} 
This task focuses on reconstructing visual images from recorded fMRI signals.
The classical methods typically transform it into 
semantic content identification~\cite{cox2003functional}, image classification~\cite{haxby2001distributed, thirion2006inverse} or retrieval tasks \cite{kay2008identifying}, or limited to simple images such as handwritten digits \cite{schoenmakers2013linear}. 
 With the development of deep networks, Generative Adversarial Networks (GANs) and VAE are employed to better reconstruct natural images \cite{shen2019deep, mozafari2020reconstructing,ren2021reconstructing, gu2022neurogen}.
Recently, diffusion models~\cite{rombach2022high} have spurred studies~\cite{chen2023seeing,ozcelik2023natural, scotti2023reconstructing,zeng2023controllable,ferrante2023brain, fang2024alleviating} in natural image decoding. Particularly, 
Chen~\etal~\cite{chen2023seeing} improve visual quality and semantic consistency of  reconstructed images  using masked brain modeling and the Latent Diffusion Model (LDM). 
However, the SD text condition utilized in their approach cannot properly handle the learning of  detailed spatial conditions.
To achieve consistent reconstructions both semantically and visually, some works \cite{ozcelik2023natural, scotti2023reconstructing} use CLIP text and image features along with VAE to generate blurry intermediate images, further refined with image-to-image Versatile Diffusion~\cite{xu2023versatile}. Other recent works employ depth maps \cite{ferrante2023brain} and silhouette maps \cite{zeng2023controllable} for low-level consistency, yet potential information gaps remain.
In contrast, NeuroPictor, our proposed approach, adapts the diffusion model to directly extract and decode fMRI signals without relying on potentially insufficient intermediate information.

\noindent
{\bf Conditional Diffusion Model.}
The vanilla stable diffusion~\cite{rombach2022high} generates images based on text prompts. 
Conditional diffusion models incorporate additional modalities like category~\cite{bar2023multidiffusion}, sketch~\cite{bashkirova2023masksketch}, depth~\cite{mou2023t2i}, normal~\cite{huang2023composer}, and semantic maps~\cite{zeng2023controllable} to guide image generation.
 Efforts have focused on fine-tuning stable diffusion for tasks such as image editing~\cite{brooks2023instructpix2pix} and personalization~\cite{gal2022image, ruiz2023dreambooth}.
T2i-adapter~\cite{mou2023t2i} modifies stable diffusion's feature maps directly for precise control, while ControlNet~\cite{zeng2023controllable} learns a trainable-copy of SD to control generation with localized image conditions. Inspired by these methods, we employ a low-Level manipulation network, combining feature transformation techniques with a U-net Encoder, enabling intricate low-level detail learning from fMRI signals, leading to a fMRI-controllable generation model.

\noindent
{\bf fMRI Representation.} 
Learning a cross-subject representation for the fMRI data could enhance the down-stream tasks. In \cite{chen2022seeing,chen2023cinematic}, the masked brain modeling is pretrained 
on the Brain, Object, Landscape Dataset (BOLD5000)~\cite{chang2019bold5000} to grasp useful context knowledge of fMRI data. However, due to the varied lengths of flattened 1-D fMRI voxels across individuals, only a single model can be trained per subject for subsequent image decoding.
Recently, Qian~\etal~\cite{qian2023fmri} present a large-scale transformer-based fMRI autoencoder 
using the UK Biobank dataset (UKB)~\cite{miller2016multimodal}, transforming individual native-space fMRI signals into unified 2D brain activation images for multi-individual brain modeling in a large-scale latent space. 
Different from previous works~\cite{chen2022seeing,qian2023fmri} learn an individual-specific fMRI-to-image decoder,
our NeuroPictor shows that leveraging the pretraining on multiple-individual fMRI-image pairs, can actually significantly improve the  fMRI-to-image decoding of one individual. 
Concurrent work~\cite{scotti2024mindeye2} also aims to develop a shared-subject fMRI-to-image model. Different from their approach, which uses subject-specific ridge regression to map individuals' fMRI data into a common space, our method directly transforms native-space fMRI signals from any individual into unified 2D activation images. This eliminates the need for subject-specific encoding modules.

\section{Method}

\begin{figure}[t]
\begin{centering}
\includegraphics[width=0.8\linewidth]{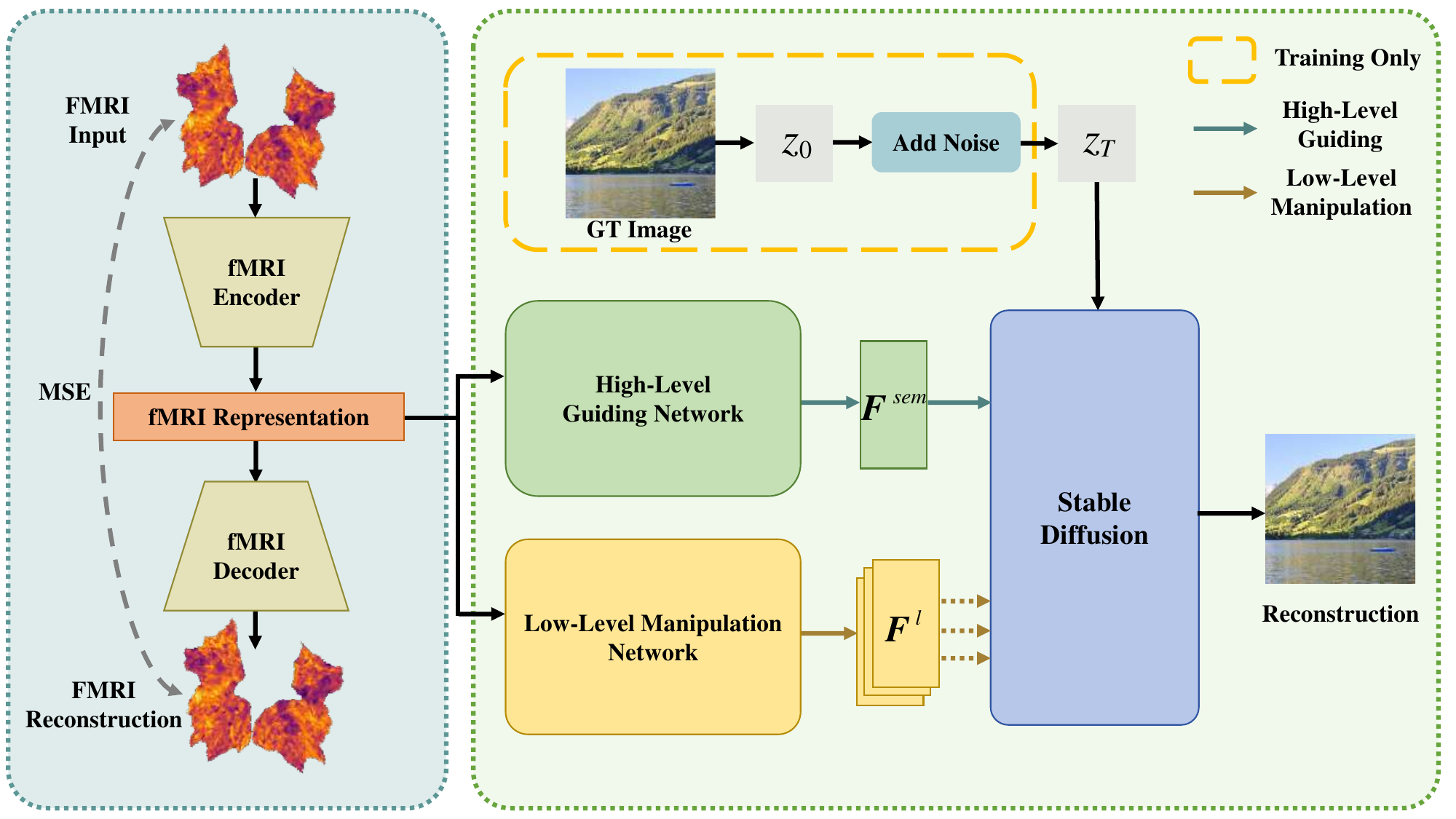}
\par\end{centering}
 \caption{ \small 
Our NeuroPictor framework is trained in three steps for fMRI-to-image decoding.
\textbf{i)} the fMRI calibrated-encoding stage, which establishes a universal latent fMRI space across multiple individuals;
\textbf{ii)} the fMRI-to-image multi-subject pre-training stage, which achieves multi-level modulation through perceptual learning. 
\textbf{iii)} the fMRI-to-image single-subject refining stage, using the same strategy in step ii but focuses on refinement on particular subject.
 \label{fig:overview}}
\end{figure}

\noindent
{\bf Overview.}\quad
The framework of NeuroPictor is illustrated in \cref{fig:overview}. 
NeuroPictor is trained by three steps of fMRI calibrated-encoding, multi-subject pre-training and single-subject refining for fMRI-to-image decoding.

\noindent\textbf{i)} In the fMRI calibrated-encoding stage, a universal latent fMRI space is learned to represent fMRI signals across multi-individuals, addressing the individual differences and data scarcity.
This latent space is trained via a transformer-based auto-encoder for reconstructing the fMRI brain surface maps (\cref{sec:encoder}). 

\noindent\textbf{ii)} In the multi-subject pre-training stage, we learn to extract high-level semantic features (\cref{sec:highlevel}) and low-level structure features (\cref{sec:lowlevel}) from the above learned fMRI latent, and then use these features as conditions to guide the generation process of diffusion model.
These components are integrated via a unified model to directly train on fMRI-to-Image decoding task on multi-subject fMRI-image pairs.

\noindent\textbf{iii)} For the particular single-subject, we use the same training strategy as in step ii but focus on the single subject to refine our NeuroPictor.

\subsection{fMRI Encoder}
\label{sec:encoder}

Inspired by the pre-processing of fMRI-PTE~\cite{qian2023fmri}, 
we transform individual fMRI signals into unified 2D brain activation images, resulting in a 1-channel image $S$ of size $256 \times 256$. 
This process unifies the fMRI representation across different individuals and thus enables training over multiple individuals to prevent missing signals of different individuals.
However, fMRI-PTE~\cite{qian2023fmri} compresses fMRI signals using VQ-GAN, contrary to our goal of learning a universal latent representation without compression. 
Instead, we directly use the masked auto-encoder to reconstruct the fMRI surface map.
In contrast to the random masking strategy of MAE, we mask all the patches in the latent space and introduce an additional \textit{guide token} along with masked tokens to reconstruct the surface map.
This forces the MAE learns to preserve all the information of fMRI signals in the \textit{guide token}, which is beneficial to learn a compressed and universal fMRI latent.
We train the autoencoder on the UKB dataset~\cite{miller2016multimodal} 
learning a universal fMRI latent space capable of encoding information from different individuals.

After training, we discard the decoder and utilize the encoder $\operatorname{E}(\cdot)$ to convert the fMRI surface map into latent representation, described by:
\begin{equation}
    \bm{S}^r = \operatorname{E}(S),
\end{equation}
where $\bm{S}^r \in \mathbb{R}^{L_r\times d_r}$, ${L_r}$ and $d_r$ represent the number of tokens and feature dimensions, respectively. 
$\bm{S}^r$ serving as the representation of fMRI data in the latent space. 
This fMRI latent is used to derive subsequent high-level semantic and low-level structural features.
Please refer to  Supplementary  for more details.

\subsection{High-Level Semantic Feature Learning}
\label{sec:highlevel}

\begin{figure}[tb]
  \centering
  \begin{subfigure}{0.49\linewidth}
    \includegraphics[width=0.95\linewidth]{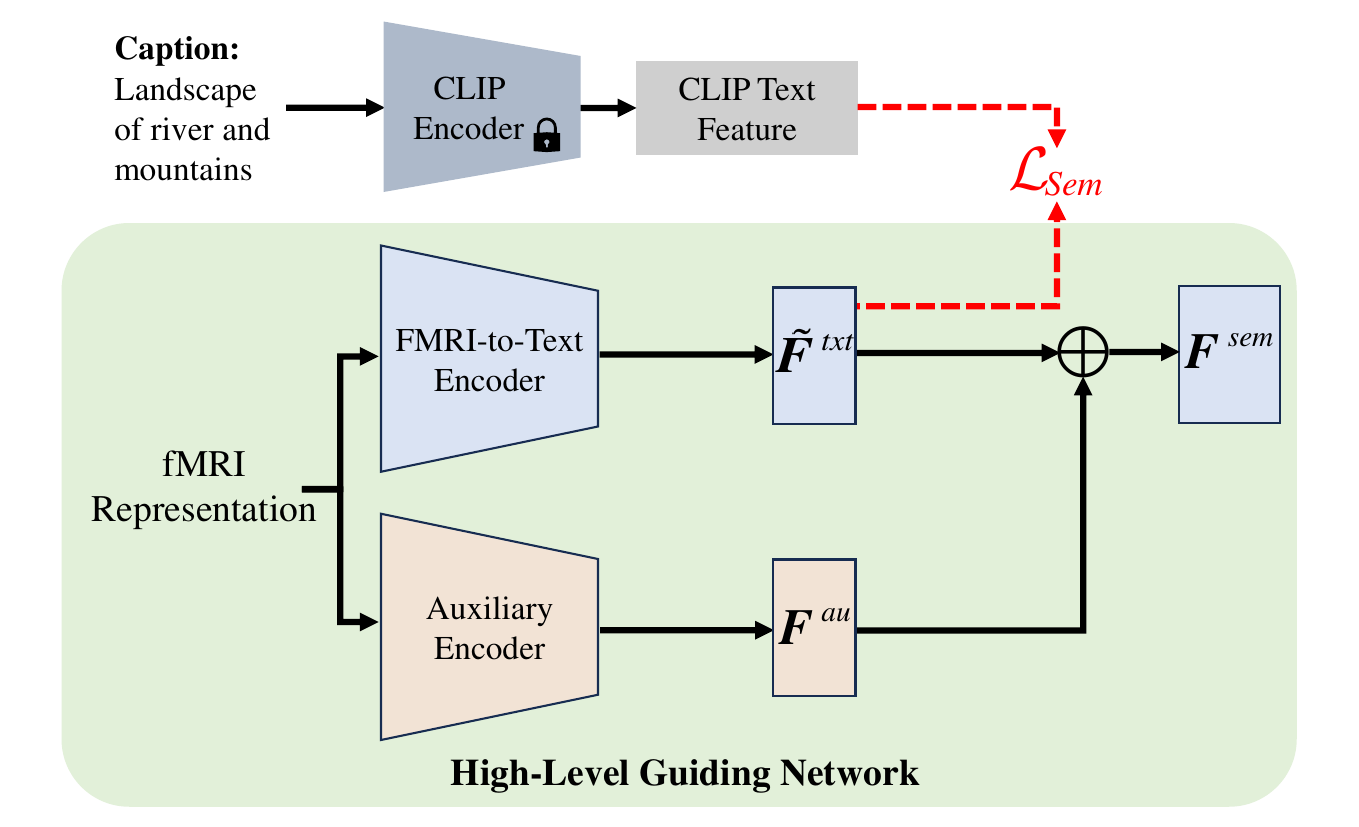}
    \caption{High-Level Guiding Network.}
    \label{fig:sem_mapper}
  \end{subfigure}
  \hfill
  \begin{subfigure}{0.49\linewidth}
    \includegraphics[width=0.95\linewidth]{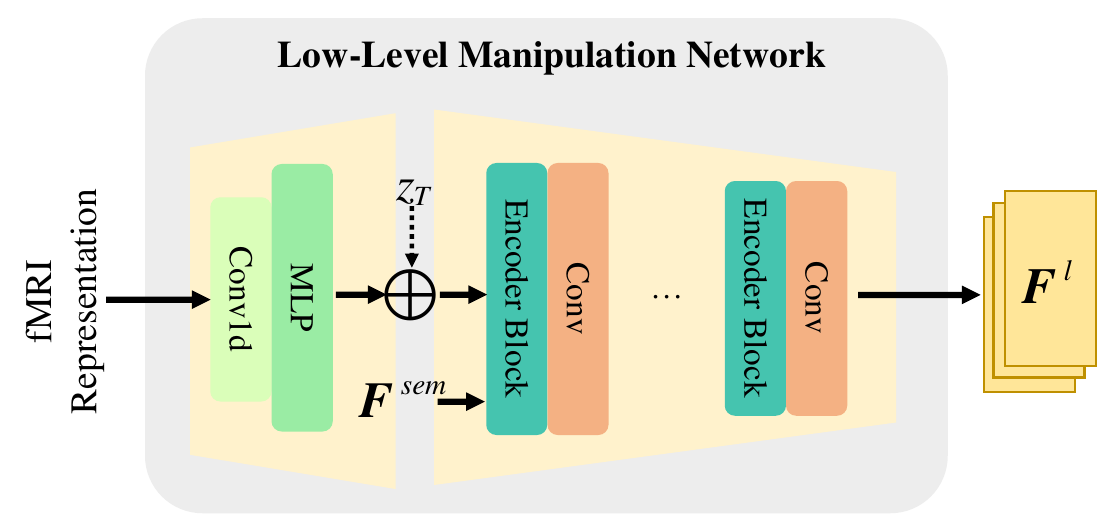}
    \caption{Low-Level Manipulation Network.}
    \label{fig:low_mapper}
  \end{subfigure}
  \caption{Framework of our High-Level Guiding Network and Low-Level Manipulation Network. }
  \label{fig:sem_mapper}
\end{figure}

The vanilla SD model uses text prompts to generate images aligned with the given text. During the text condition stage, it utilizes a fixed CLIP encoder to encode the text prompts as $\bm{F}^{txt} \in \mathbb{R}^{L_T\times d_T}$. These prompts then guide the cross-attention layers in image generation. To maximize the potential of the high-level adapted cross-attention layers, we suggest substituting text conditions with high-level semantic fMRI features.

To fully upscale SD as a fMRI-to-image model, we replace the text encoder with a semantic feature learned from fMRI signals.  Using the encoded fMRI feature $\bm{S}^r \in \mathbb{R}^{L_r\times d_r}$, we design a high-level guiding network to connect the fMRI latent space and text condition space.   Illustrated in \cref{fig:sem_mapper}, the high-level guiding network processes fMRI representations through two parallel branches: the fMRI-to-text encoder aligns fMRI signals with CLIP Text features $\bm{F}^{txt}$ of image captions, while the auxiliary encoder provides additional semantic information to compensate for inaccuracies in the fMRI-to-text encoder and capture rich semantic details from images not explicitly described in captions.

In the fMRI-to-text encoder, two 1-D convolution layers are used to downsample the token number ${L_r}$ of fMRI representation $\bm{S}^r \in \mathbb{R}^{L_r\times d_r}$. Subsequently, MLP is employed to project it into  text condition space: 
\begin{equation}
     \bm{\Tilde{F}}^{txt} = \operatorname{E}_{txt}(\bm{S}^r),
\end{equation}
where $\bm{\Tilde{F}}^{txt} \in \mathbb{R}^{L_T\times d_T}$ is used to estimate the CLIP text feature of the ground truth image caption and $\operatorname{E}_{txt}$ is the fMRI-to-text encoder we introduced. 

To further enforce the learned feature $\bm{\Tilde{F}}^{txt}$ aligned with  text condition, we use the caption of the original image to optimize it. Specifically, a frozen CLIP text encoder is used to exact text feature $\bm{F}^{txt} \in \mathbb{R}^{L_T\times d_T}$ as the ground truth text condition. And we use mean square error as the semantic loss to optimize it, which can be formulated as:
\begin{equation}
\mathcal{L}_{sem} = \frac{1}{L_T} \sum_{i=1}^{L_T} \| \bm{F}^{txt}_{(i)} - \bm{\Tilde{F}}^{txt}_{(i)} \|_2^2, 
\end{equation}
where $\bm{F}^{txt}_{(i)}$ and $\bm{\Tilde{F}}^{txt}_{(i)}$ represents the i-th token feature of $\bm{F}^{txt}$ and $\bm{\Tilde{F}}^{txt}$, respectively.

However, directly using the hard-mapped feature $\bm{\Tilde{F}}^{txt}$ is not optimal. To address inaccurate semantic information learned in the fMRI-to-text branch, we introduce an auxiliary encoder to obtain an auxiliary semantic code $\bm{F}^{au}$. The architecture of the auxiliary encoder resembles that of the fMRI-to-text encoder, with the addition of a 1D zero convolutional layer in the final layer to initialize $\bm{F}^{au}$ as a zero tensor. This auxiliary semantic code is then residually added to the text code, resulting in the semantic representation $\bm{F}^{sem}$.
\begin{equation}
\bm{F}^{sem} = \bm{\Tilde{F}}^{txt} + \bm{F}^{au}. 
\end{equation}

This adapted semantic feature $\bm{F}^{sem}$ is then fed into SD to guide the cross-attention layers.

\subsection{Low-Level Manipulation Network}
\label{sec:lowlevel}

While learning general semantic information by mapping fMRI signals into a textual condition space is feasible, bridging the gap between fMRI and intricate images, rich in low-level details, poses challenges. To enhance low-level structure guidance, we introduce a Low-Level Manipulation Network (LLMN). This network injects low-level fMRI conditions by directly manipulating feature maps in the SD's U-net, adapting the generative model to fMRI signals.

Our LLMN consists of two components:  a feature transformation module to convert fMRI representation to SD latent space, and a manipulator to incrementally trained to manipulate the SD latent.

\noindent
{\bf Feature Transformation.} 
Given an image of size $H \times W \times 3$, Stable Diffusion apply an Autoencoder similar to 
VQGAN~\cite{Esser2021taming} to convert the pixel-space image into latent space as $z_0 \in \mathbb{R}^{c \times h \times w}$. Then, a U-net~\cite{ronneberger2015u} is used to progressively denoise images in the latent space.

To directly manipulate feature maps in SD's U-net, we introduce a feature transformation technique that simultaneously performs channel-wise and dimensional-wise feature learning on fMRI embeddings \footnote{ {\small We use the terms "channel" and "dimension" to refer to the second and third dimensions of the fMRI latent tensor. }}. This connects fMRI representation learning with feature map learning, facilitating the subsequent manipulation process. Specifically, we employ two layers of 1D convolution to handle channel-wise correlations of fMRI embeddings and the MLP to conduct dimensional-wise feature learning. Finally, we use a reshape operation to form a feature map $\bm{F}^{l}_0 \in \mathbb{R}^{c \times h \times w}$.
\begin{equation}
     \bm{F}^{l}_0= \operatorname{E}_{ft}(\bm{S}^r).
\end{equation}
The feature map $\bm{F}^{l}_0$ is then fed into the subsequent U-net encoder blocks, acting as condition controlling the generation process.

\noindent
{\bf Low-Level Manipulation.} 
To regulate the generation results with fine-grained low-level details, inspired by~\cite{zhang2023adding}, we incrementally fine-tune the SD to improve the coherence between fMRI signals and visual stimulus.
Utilizing a residual connection network, we effectively bridge the gap between fMRI representations and real natural images. 
Specifically, we incrementally manipulate the blocks of SD U-Net by adding low-level fMRI conditions within the proposed manipulation network.
As depicted in \cref{fig:overview}, our manipulation network additionally take as inputs the feature map $\bm{F}^{l}_0$ converted from the fMRI signal.
Then, to manipulate the SD latent in a fine-grained manner. 
We utilize a series of transformer layers to manipulate different blocks of SD.
Denoted as $\operatorname{E_{U}}$, this series of feature maps are produced by:
\begin{equation}
     \bm{F}^{l}=\operatorname{E_{U}}(\bm{F}^{l}_0), 
\end{equation}
where $\bm{F}^{l} = \{\bm{F}^{l}_{(i)}|i=1,\cdots,13\}$ and $\bm{F}^{l}_{(i)}$ is the i-th feature map produced by i-th encoder block. 

These feature maps are passed into zero convolution layers and residually added with the outputs of the middle block and decoder blocks of locked SD branch as,
\begin{equation}
\label{eqn:control}
     \tilde{\bm{F}^{l}}=\bm{F}_{sd} + \alpha \mathcal{Z}(\bm{F}^{l}),
\end{equation}
where $\mathcal{Z}$ represent the zero convolution layer, $\bm{F}_{sd}$ represent the latent codes in SD U-net, $\alpha$ is a hyperparameters that balance the high-level semantic guidance with low-level detail manipulation. 
In this way, the feature maps in SD U-net is updated.
We pass the feature map $\tilde{\bm{F}^{l}}$ into the subsequent layers of locked SD model to control the generation process. As a result, with the aid of the SD decoder, we obtain the final reconstructed image corresponding to specific fMRI signals. 
This incremental fine-tuning strategy is more efficient and stable than fully fine-tuning the SD.

\subsection{Training and Inference}

\noindent
{\bf Training Objectives.} 
During the training process, our NeuroPictor  takes fMRI-image pairs as inputs. The input natural image $\bm{X}$ of size $512 \times 512$ is firstly compressed into latent space as $\bm{z}_0 \in \mathbb{R}^{64 \times 64 \times 4}$. Then, the diffusion process produces a noisy version $\bm{z}_t$ of $\bm{z}_0$ by progressively adding noise for $t$ time steps to the initial feature $\bm{z}_0$. In the denoising stage, we use a frozen U-Net in Stable Diffusion and a trainable LLMN to jointly predict a denoised variant of the noisy input $\bm{z}_t$, conditioning on time step $\bm{t}$, semantic condition $\bm{F}^{sem}$, and a low-level feature map $\bm{F}^{l}$ converted from the fMRI representaion $\bm{S}^r$. The learning objective of the latent denoising stage can be formulated as:
\begin{equation}
    \mathcal{L}_{dif} = \mathbb{E}_{\bm{z}_0, \bm{t}, \bm{S}^r, \epsilon \sim \mathcal{N}(0, 1) }\Big[ \Vert \epsilon - \epsilon_\theta(\bm{z}_{t}, \bm{t}, \bm{S}^r, \bm{F}^{sem})) \Vert_{2}^{2}\Big],
\label{eq:difloss}
\end{equation}
where $\epsilon_\theta$ represent the the LLMN. For simplify, we do not include the parameters of fMRI encoder $\operatorname{E}$ and semantic mapper $\operatorname{E}_T$ in \cref{eq:difloss}. But these models are all trainable, and the parameters will be optimized with backpropagation of the gradients of $\bm{S}^r$ and $\bm{F}^{sem}$.

We combine diffusion loss with semantic loss (\cref{sec:highlevel}) to guide the training process. The final objective function is: 
\begin{equation}
\label{eq:overloss}
\mathcal{L}=\mathcal{L}_{dif} + \lambda \mathcal{L}_{sem},
\end{equation}
where $\lambda$ denotes the loss weight to balance both terms.

\noindent
{\bf Training Strategy.}
Benefiting from the unified fMRI representation in \cref{sec:encoder}, NeuroPictor can be trained using fMRI-image pairs from different individuals. Thus, we split the training into three stages: 
i) \textit{fMRI calibrated-encoding}: We pre-train the calibrated-encoder to enable the cross-subject fMRI latent space;
ii) \textit{Multi-subject pre-training}: During pretraining, the Stable Diffusion branch and LLMN are initialized with Stable Diffusion v2.1 trained on a large-scale dataset. We jointly train the full model, excluding the auxiliary encoder, using about 67,000 fMRI-image pairs from 8 subjects in the Natural Scenes Dataset~\cite{allen2022massive} for pretraining, conducting 100k iterations to obtain a generalized fMRI-to-image generation model for these 8 subjects. 
iii) \textit{single-individual refining}: for better modeling individually independent perception patterns, we fine-tune the network for an additional 60k iterations.

\noindent
{\bf Inference.}
To maintain visual consistency, we balance high-level semantic guidance and low-level detail manipulation using a control scale instead of integrating them too faithfully. 
Essentally, focusing too much on high-level aspects can result in lost image detail, while prioritizing low-level features may impact the semantic processing abilities of the SD model, causing generated images to prioritize structural accuracy over semantic consistency. 
Thus in the inference stage, we can adjust control scale $\alpha$ to weight the SD output and LLMN output in \cref{eqn:control}.
We show how control scale influence the generation process in \cref{fig:teaser}. As the control scale enlarges from left to right, the decoded images gradually achieve the reconstruction of visual stimuli, progressing from high-level semantics only to high-level semantics and low-level structures consistently.

\section{Experiments}

\begin{figure}[t] \small
\begin{centering}
\includegraphics[width=0.97\linewidth]{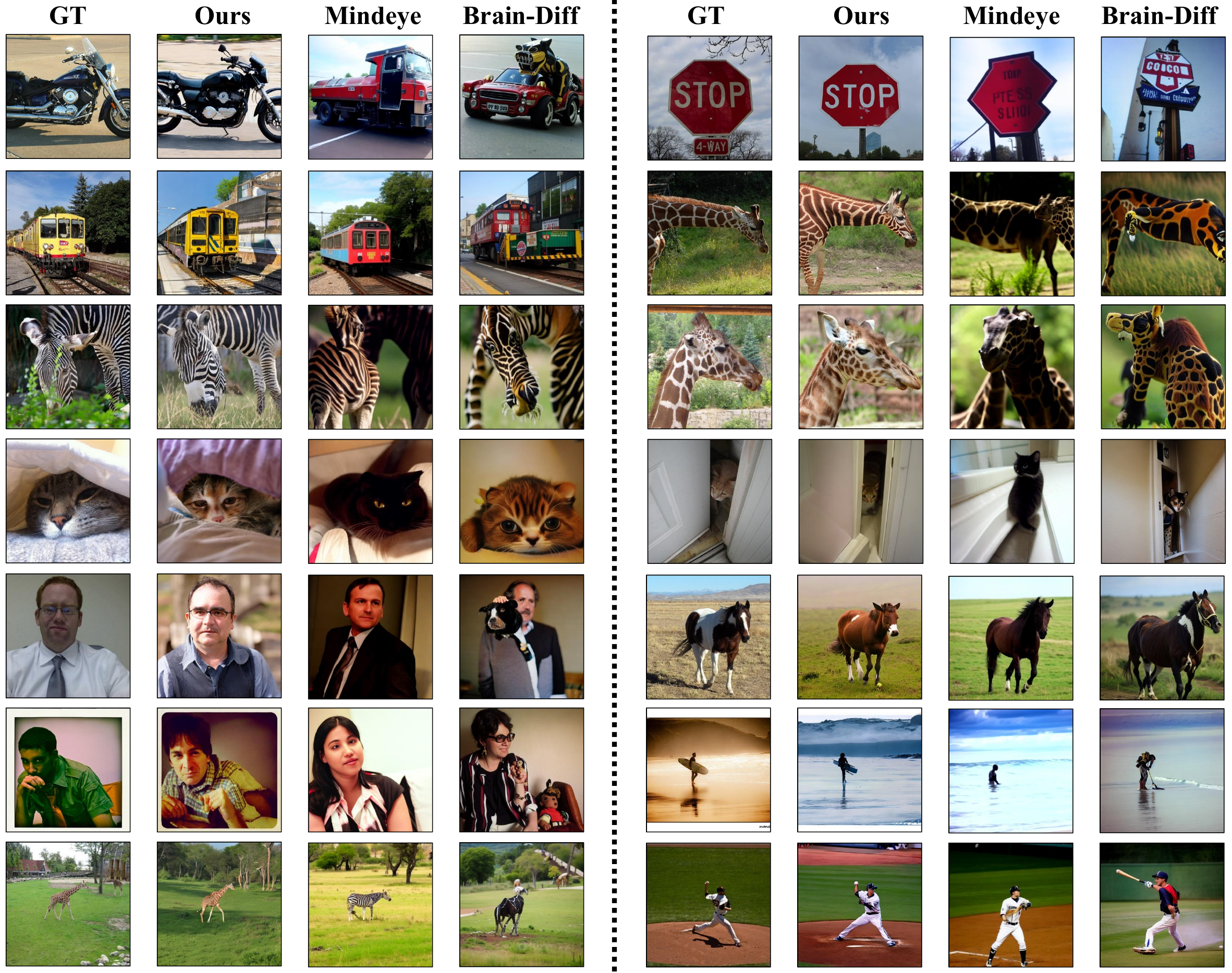}
\par\end{centering}
 \caption{ \small Qualitative comparision of our NeuroPictor  and previous state-of-the-art methods. 
Compared with other methods, our~NeuroPictor  achieves both high-level semantics and low-level structures consistency. \label{fig:main_results}}
\end{figure}

\subsection{Experimental setup}

\noindent
{\bf Dataset.} 
We use Natural Scenes Dataset (NSD) \cite{allen2022massive} for both training and evaluation. NSD contains visual image stimulus and corresponding fMRI recordings of 8 subjects, with each subject viewing 8,000-9,000 images. The original images are collected from MS-COCO dataset \cite{lin2014microsoft}, which are consisted of complex natural images. Following \cite{ozcelik2023brain}, we use the corresponding captions of the images in COCO dataset for training. For both training and evaluation, we average three trials of fMRI signal of the same images. 
Previous works only use the data of subjects 1, 2, 5 or 7 to train a model for each subject. Because these subjects complete all scanning sessions, sharing the same $982$ images as testing data. 
As our unified fMRI representations enable training across different person, we use all training data of 8 subjects to conduct a pretraining for modeling shared activity of all subjects. This scales up our training data to about 67k pairs. Then, following the setup of previous work, we finetune our NeuroPictor  using the train split of subject 1, 2, 5 and 7, respectively, and evaluate our model using the shared test split.

\noindent
{\bf Evaluation Metrics.} 
We follow the metrics of Mindeye \cite{scotti2023reconstructing} to evaluate both high-level and low-level consistency. On the low-level aspect, we use pixelwise correlation, Structural Similarity Index Metric (SSIM) \cite{wang2004image}, AlexNet(2), and AlexNet(5). High-level metrics are calculated by extracting features using specific networks, including EffNet-B \cite{tan2019efficientnet}, SwAV \cite{caron2020unsupervised}, Inception \cite{szegedy2016rethinking}, and CLIP \cite{radford2021learning}. Please refer to Supplementary for more details.

\noindent
{\bf Implementation Details.} 
The pretraining experiments are conducted using 6 NVIDIA RTX A6000 GPUs, with a batch size set to 96. During this stage, we exclude the auxiliary encoder and semantic loss to learn a base model. The entire network is trained for 100,000 iterations to achieve convergence. In the finetuning stage, 2 NVIDIA RTX A6000 GPUs are employed to finetune the model for each subject, spanning 60,000 iterations. The batch size for finetuning is set to 32. In \cref{eq:overloss}, the parameter $\lambda$ is set to $0.1$. Additionally, we enables NeuroPictor to adapt to Classifier-Free Guidance~\cite{ho2022classifier} by randomly replacing 5\% of the semantic features with unconditional CLIP embeddings corresponding to empty characters in the end of each stage. During evaluation, the unconditional guidance scale is set to 5.0. 

\subsection{Main Results}

\begin{table}[!t] 
    \centering
    \renewcommand\arraystretch{1.0}
     \caption{\small Quantitative comparison of within-subject brain decoding of our NeuroPictor  and the previous state-of-the-art mothods on Natural Scenes Dataset. }
    \label{tab:recon_eval}
    \scalebox{0.78}{
    \begin{tabular}[width=0.97\linewidth]{lcccccccc}
        \toprule
        \multicolumn{1}{c}{\multirow{2}{*}{\textsc{Method}}} & \multicolumn{4}{c}{Low-Level} & \multicolumn{4}{c}{High-Level} \\
        \cmidrule(l){2-5} \cmidrule(l){6-9}
        & PixCorr $\uparrow$ & SSIM $\uparrow$ & AlexNet(2) $\uparrow$ & AlexNet(5) $\uparrow$ & Inception $\uparrow$ & CLIP $\uparrow$ & EffNet-B $\downarrow$ & SwAV $\downarrow$  \\
        \midrule
        Lin et al.~\cite{lin2022mind} & $-$ & $-$ & $-$ & $-$  & $78.2\%$ & $-$ & $-$ & $-$ \\
        Takagi...~\cite{takagi2023high} & $-$ & $-$ &  $83.0\%$ & $83.0\%$  & $76.0\%$ & $77.0\%$ & $-$ & $-$ \\
        Gu et al.~\cite{gu2022decoding} & $.150$& $.325$ & $-$ & $-$  & $-$ & $-$ & $.862$ & $.465$ \\
        Brain-Cap~\cite{ferrante2023brain} & $\mathbf{.353}$ & ${.327}$ & ${89.0\%}$ & ${97.0\%}$  & ${84.0\%}$ & ${90.0\%}$ & $-$ & $-$ \\
        Brain-Diff~\cite{ozcelik2023brain} & ${.254}$ & ${.356}$ & ${94.2\%}$ & ${96.2\%}$  & ${87.2\%}$ & ${91.5\%}$ & ${.775}$ & ${.423}$ \\
        MindEye~\cite{scotti2023reconstructing} & ${.309}$ & ${.323}$ & ${94.7\%}$ & ${97.8\%}$  & ${93.8\%}$ & $\mathbf{94.1\%}$ & ${.645}$ & $.367$ \\
        \midrule
        NeuroPictor (w/o ft) & $.141$ & ${.349}$ & ${91.4\%}$ & ${95.7\%}$  & ${88.3\%}$ & ${88.9\%}$ & ${.722}$ & ${.417}$ \\
        NeuroPictor  & $.229$ & $\mathbf{.375}$ & $\mathbf{96.5\%}$ & $\mathbf{98.4\%}$  & $\mathbf{94.5\%}$ & ${93.3\%}$ & $\mathbf{.639}$ & $\mathbf{.350}$ \\
        \bottomrule
    \end{tabular}
    }
\end{table}

\begin{table*}[!t]
    \centering
    \captionsetup{font=small}
    \renewcommand\arraystretch{1.0}
    \setlength{\tabcolsep}{1pt}
    \small
     \caption{ \small 
Ablation study on Subject-1. "Random init" means the fMRI Encoder starts without pre-trained weights from \cref{sec:encoder}. Instead, it's trained from scratch using a standard Xavier uniform initialization. "Frozen" means the fMRI Encoder remains unchanged during fMRI-to-image training. "Full" initializes the fMRI Encoder with pre-trained weights, making it trainable during fMRI-to-image training. "Multi Pret." indicates whether fMRI-to-image multi-subject pre-training (subjects 1-8) is done. If not, the entire network is trained directly on Subject-1 data. 
    }
    \scalebox{0.7}
    {\begin{tabular}{lcccccccccc}
        \toprule
        \multicolumn{3}{c}{Method}
        & \multicolumn{4}{c}{Low-Level} & \multicolumn{4}{c}{High-Level} \\
        \cmidrule(l){1-3} \cmidrule(l){4-7} \cmidrule(l){8-11}
        Model & fMRI Enc. & Multi Pret. & PixCorr $\uparrow$ & SSIM $\uparrow$ & AlexNet(2) $\uparrow$ & AlexNet(5) $\uparrow$ & Inception $\uparrow$ & CLIP $\uparrow$ & EffNet-B $\downarrow$ & SwAV $\downarrow$  \\
        \midrule
        \modelname & random init & & $.128$& $.323$ & $91.2\%$ & $95.1\%$  & $84.9\%$ & $86.0\%$ & $.769$ & $.448$ \\
        \modelname & [CLS] & & $.139$& $.313$ & $90.8\%$ & $94.2\%$  & $87.9\%$ & $87.6\%$ & $.736$ & $.437$ \\
        \modelname & frozen & & $.220$& $.342$ & $97.5\%$ & $98.8\%$  & $93.8\%$ & $92.8\%$ & $.657$ & $.360$ \\
        \modelname & full &  & $.265$& $.368$ & $98.2\%$ & $99.1\%$  & $94.7\%$ & $93.4\%$ & $.641$ & $.342$ \\
        \midrule
        W/o Finetune & full & \checkmark & $.167$ & $.350$ & $95.1\%$ & $97.7\%$ & $89.1\%$ & $90.7\%$ & $.698$ & $.399$ \\
        W/o LLMN & full & \checkmark & $.111$& $.307$ & $75.4\%$ & $88.4\%$  & ${89.4\%}$ & ${89.9\%}$ & ${.746}$ & ${.460}$ \\
        W/o Au. Enc. & full & \checkmark & $.221$& $.276$ & $91.1\%$ & $97.5\%$  & $94.5\%$ & $91.0\%$ & $.668$ & $.387$ \\
        W/o $\mathcal{L}_{sem}$ & full & \checkmark & $.255$& $.373$ & $98.0\%$ & $98.9\%$  & $94.3\%$ & $93.0\%$ & $.647$ & $.350$ \\
        \modelname & full & \checkmark & $\mathbf{.277}$ & $\mathbf{.385}$ & $\mathbf{98.8\%}$ & $\mathbf{99.3\%}$  & $\mathbf{96.2\%}$ & $\mathbf{94.5\%}$ & $\mathbf{.619}$ & $\mathbf{.334}$  \\
        \bottomrule
    \end{tabular}}
    \label{tab:ablation}
\end{table*}

In this section, we comprehensively evaluate NeuroPictor against existing methods, using both quantitative and qualitative comparisons.

\noindent \textbf{Competitors.}  We use the same test set as \cite{scotti2023reconstructing} and follow the values reported in \cite{scotti2023reconstructing}. Additionally, we include a new method \cite{ferrante2023brain} for comparision, which is also trained and tested on NSD. 

\noindent \textbf{Quantitative results. }
The results of quantitative comparison of our NeuroPictor  and previous state-of-the-art methods are displayed in \cref{tab:recon_eval}. We generate one sample for each subject and calculate the low-level and high-level metrics following the official code of \cite{scotti2023reconstructing}. We average the metrics across 4 subjects (\ie Subject-1, 2, 5 and 7) and report both the pre-trained (\ie ``w/o ft'' in line 7 of \cref{tab:recon_eval}) and fine-tuned results. The results show that our NeuroPictor  outperforms other methods in six out of eight metrics, demonstrating that our reconstruction results can maintain semantic consistency while preserving low-level details present in the original images. 
We find that our method falls behind in pixel correlation metrics compared to three other models. This could potentially be attributed to the fact that other methods employ an image-to-image diffusion model and use a blurred image produced by VAE as an initialization. As reported in \cite{ren2021reconstructing, ferrante2023brain}, such a blurred image inherently exhibits higher pixel correlation than finely reconstructed results. Furthermore, our non-finetuned model achieves results comparable to other models, except for Mindeye. This validates that our training on multi-individual data effectively captures shared perceptual features across different individuals.

\noindent \textbf{Qualitative results. }
We conduct qualitative comparisons with previous state-of-the-art works, namely Mindeye and Brain-diffuser, through visualization. We utilize the official codes to generate images for comparison. As shown in \cref{fig:main_results}, our reconstruction results exhibit high visual consistency with the ground truth images, particularly in capturing some of the underlying details such as object structures, positions, and human poses. 
For instance, the appearance and poses of black motorcycles and trains, the positions of animals, as well as the gesture of people holding skateboards or engaged in throwing actions. 
We successfully reconstruct many challenging cases, such as a red "STOP" sign (row 1), animals like zebras and giraffes showing only their heads (row 3), and cats peeking out from under blankets or through door gaps (row 4). 
Additionally, our NeuroPictor successfully decodes the white borders (row 6) in visual stimuli, while other methods fails, highlighting the effectiveness of our direct low-level manipulation.
Our reconstruction results maintain certain local matches with the original images, attributed to our end-to-end training fashion and pretraining on fMRI-image pairs.

\subsection{Ablation Study}
\label{sec:ablation}

\begin{figure}[t] \small
\begin{centering}
\includegraphics[width=0.75\linewidth]{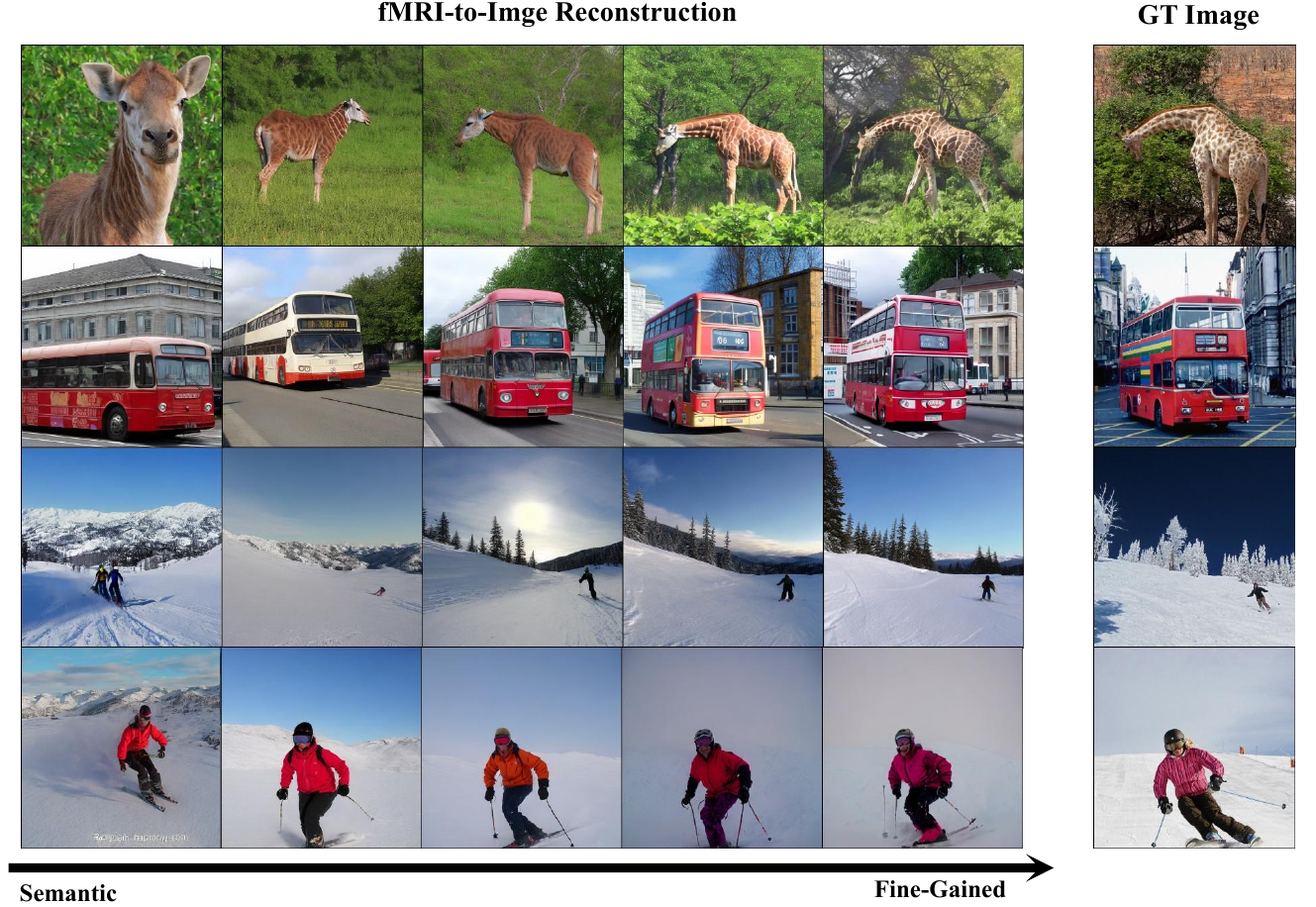}
\par\end{centering}
 \caption{ \small Interpolating the control scale between 0 and 1 transitions the reconstructed image from semantic consistency to fine-grained control.
\label{fig:controlscale}}
\end{figure}

In this section, we perform ablation studies to evaluate the effects of our NeuroPictor framework techniques, focusing on Subject-1 for efficiency. Results are shown in \cref{tab:ablation}, with additional discussion on failure cases in the Supplementary.

\noindent
{\bf Ablation of fMRI Encoder.} 
We investigate the fMRI encoder's pretraining and feature selection impacts. Compared with training from scratch using a Xavier uniform initialization, pretraining an autoencoder on the UKB dataset before starting fMRI-to-image reconstruction significantly enhances performance, as shown in \cref{tab:ablation}. Furthermore, fine-tuning the pretrained fMRI encoder, instead of freezing it during the fMRI-to-image stage, leads to improvements in all metrics. Switching the fMRI encoder's output from a high-dimensional feature to a [CLS] token results in reduced reconstruction quality.

\noindent
{\bf  Ablation of Multi-Subject Pre-training.}
We validate the impact of both our pretraining and finetuning in fMRI-to-image reconstruction stage by removing one at a time. The results indicate that pretraining on 8 subjects and finetuning on a single subject simultaneously improve the model's performance in both high-level and low-level aspects. 

\noindent
{\bf Ablation of High-Level Guiding Network and Low-Level Manipulation Network.}
Testing the High-Level Guiding Network without semantic loss shows its importance for achieving high-level consistency in reconstructions. Removing the auxiliary encoder and relying solely on the fMRI-to-Text Encoder for alignment with CLIP text features significantly degrades performance, highlighting the challenge of bridging semantic gaps between text captions and visual stimuli. Additionally, the presence or absence of the Low-Level Manipulation Network (LLMN) indicates its primary function in refining details, while also secondarily aiding in the preservation of high-level semantic coherence. As shown in \cref{fig:controlscale}, we also visualize generated results with a control scale varying between 0 and 1 to confirm the efficacy of both low-level and high-level pipelines in NeuroPictor.

\section{Conclusions}

In conclusion, this paper introduces NeuroPictor, a novel framework that tackle fMRI-to-image task by directly modulating the generation process of diffusion models using fMRI signals. Unlike previous approaches that focus on associating fMRI signals with pre-trained diffusion model conditions, NeuroPictor offers detailed control over image creation by dividing the fMRI-to-image process into three key steps: fMRI calibrated-encoding, fMRI-to-image single-subject refining and fMRI-to-image single-subject refining.
NeuroPictor learns to exact high-level semantic features from fMRI signals and incrementally fine-tunes the diffusion model with a low-level manipulation network, providing precise structural instructions. Through training on over 60,000 fMRI-image pairs from diverse individuals, our model demonstrates superior fMRI-to-image decoding capabilities, particularly in within-subject scenarios.

\section*{Acknowledgements}
Jingyang Huo, Yun Wang and Jianfeng Feng are with Institute of Science and Technology for Brain-inspired Intelligence, Fudan University. 
Chong Li is with School of Computer Science, Fudan University. 
Yikai Wang, Xuelin Qian and Yanwei Fu are with School of Data Science, Fudan University. Xuelin Qian is now with Northwestern Polytechnical University.
The computations in this research were performed using the CFFF platform of Fudan University.


\bibliographystyle{splncs04}
\bibliography{main}

\clearpage
\noindent \textbf{\Large Supplementary Material}
\vspace{0.1in}
\setcounter{section}{0}


\section{Implementation Details}

\subsection{Cortex used for fMRI decoding} 
\begin{figure*}[t]
\begin{centering}
\includegraphics[width=0.4\linewidth]{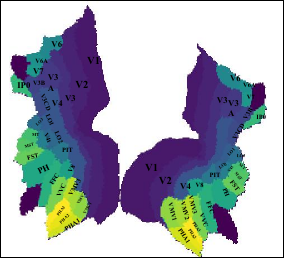}
\par\end{centering}
 \caption{Visualization of VC RoIs. \label{fig:vc_roi}}
\end{figure*}

We use the same cortex as fMRI-PTE \cite{qian2023fmri}. Specifically, early and higher VC RoIs selected from the HCP-MMP atlas in fsLR32K surface space are shown in \cref{fig:vc_roi}, including ``V1, V2, V3, V3A, V3B, V3CD, V4, LO1, LO2, LO3, PIT, V4t, V6, V6A, V7, V8, PH, FFC, IP0, MT, MST, FST, VVC, VMV1, VMV2, VMV3, PHA1, PHA2, PHA3''.

\subsection{More Training Details}

For both pretraining and finetuning, we initially use the AdamW optimizer with a learning rate of $10^{-4}$ and a weight decay of 0.01. Additionally, to enable NeuroPictor to adapt to Classifier-Free Guidance~\cite{ho2022classifier}, we randomly replace 5\% of the semantic features with unconditional CLIP embeddings corresponding to empty characters at the end of each stage. During the stage where we replace 5\% of the semantic features, the learning rate is set to $10^{-5}$, with 12,000 iterations for multi-subject pretraining and 5,000 iterations for single-subject finetuning. Stable Diffusion v2.1 serves as the initialization for the SD branch.

\subsection{Metric Details}

On the low level aspect, four metrics are introduced, including pixelwise correlation, Structural Similarity Index Metric (SSIM) \cite{wang2004image}, AlexNet(2) and AlexNet(5). Pixelwise correlation and SSIM are calculated by average the correlation distance of every paired grond truth image and reconstruction, while AlexNet(2) and AlexNet(5) represent the accuracy of two-way classification using the feature of the corresponding layers of AlexNet. High level metrics are calculated by extracting the feature using specific networks. Similarly,  EffNet-B and SwAV is obtained by calculating the average distance of the features exacted by EfficientNet-B1 \cite{tan2019efficientnet} and SwAV-ResNet50 \cite{caron2020unsupervised}. The metric for Inception \cite{szegedy2016rethinking} and CLIP \cite{radford2021learning} represent the accuracy of two-way classification based on corresponding feature.

\subsection{Details on fMRI Encoder}
\label{sec:fmrienc}

In this section, we will provide a detailed explanation of the specific architecture and training strategy used to obtain the autoencoder for the fMRI encoder.

\noindent
{\bf Model Architectures.} 
We employ an autoencoder structure \cite{qian2023semantic}, similar to MAE \cite{he2022masked}, for the reconstruction of fMRI surface maps $S$. Specifically, we process the original $256 \times 256$ fMRI surface map using a patch size of 16. Each patch is tokenized into a 1024-dimensional embedding. This embedding is then combined with positional encoding and passed through subsequent transformer blocks. For the transformer, we set the depth to 24 and the number of heads to 16. After layer normalization, the output features, denoted as $\bm{S}^r \in \mathbb{R}^{257\times 1024}$, are obtained as the fMRI representation used in our study.

Next, we introduce the architecture of the decoder. We take the [CLS] token from $\bm{S}^r$ and pass it through a fully connected layer, replacing other patch tokens with a learnable \textit{guide token}. Then, these features are passed into the transformer blocks of the decoder, which possess a structure symmetric to that of the encoder, with a depth of 24 and 16 heads. Finally, a fully connected layer followed by an ``unpatchify" operation is utilized to map the latent features to the pixel space of the original fMRI surface map. We denote the reconstructed fMRI surface map as $\tilde{S}$.

\noindent
{\bf Training.} 
The training objective of this autoencoder is to reconstruct the original fMRI surface map. We use mean squared error (MSE) as the measure for the reconstruction loss, formulated as:
\begin{equation}
\label{eq:rec}
\mathcal{L}_{\text{rec}} = | S - \tilde{S} |_2^{2},
\end{equation}
where $S$ and $\tilde{S}$ represent the ground truth and predicted fMRI surface map.

In fMRI calibrated-encoding stage, we train the autoencoder on the UKB dataset \cite{miller2016multimodal} for 120K iterations with 8 NVIDIA A100 GPUs. Following \cite{qian2023fmri}, we use rest-fMRI of 40k subjects of UKB dataset to tackle multi-individual pre-training task. The batch size is set to 384, and the AdamW optimizer is used with a linear learning rate schedule. The learning rate is initialized to 0.0005. With this training on a large-scale dataset, we learn a universal fMRI latent space capable of encoding information from different individuals. Then we remove the decoder part and use the encoder to map the original fMRI signal into latent representations for downstream fMRI-to-image reconstruction.

\subsection{Details on Low-Level Manipulation Network}
\label{sec:lowlevel}
We design a Low-Level Manipulation Network (LLMN) to manipulate the SD latent. This network comprises a feature transformation module and a manipulator.

To start, the fMRI representation obtained in \cref{sec:fmrienc} serves as input for the LLMN. Initially, we introduce a feature transformation technique that links fMRI representation learning with feature map learning. This involves employing two layers of 1D convolution to address the channel-wise correlation of fMRI embeddings and using an MLP for dimensional-wise feature learning. Subsequently, a reshape operation is applied to create a feature map $\bm{F}^{l}_0 \in \mathbb{R}^{320 \times 64 \times 64}$. 
Next, we set up a trainable copy of the U-Net encoder in SD as a manipulator. This manipulator mirrors the architecture of the SD U-Net encoder, comprising 12 encoder blocks and 1 middle block that includes ResBlock, DownSample, and SpatialTransformer. The manipulator primarily takes the feature map $\bm{F}^{l}_0$ derived from the fMRI signal as input. Additionally, we utilize the semantic representation $\bm{F}^{sem}$ as input for the SpatialTransformer to execute a cross-attention operation with the feature map. The feature map $\bm{F}^{l}_0$ is added residually with the noisy latent $z_T$ and then inputted into the subsequent 13 encoder blocks to generate a series of feature maps $\bm{F}^{l} = \{\bm{F}^{l}_{(i)}|i=1,\cdots,13\}$. 
Finally, these feature maps are subsequently fed into zero convolution layers and added residually with the outputs of the middle block and decoder blocks of the locked SD branch. This process updates the feature maps in the SD U-Net, facilitating direct low-level manipulation in the generation process using fMRI signals.

Different from~\cite{zhang2023adding}, which process specific images as condition and utilize conv2d operations to handle spatial information, we simultaneously conduct channel-wise and dimensional-wise feature learning on fMRI representations. This approach bridges the gap between fMRI representation learning and feature map learning, capturing rich information embedded in fMRI representations. Furthermore, we finetune the full model by jointly training high-level guiding networks and low-level manipulation networks without detaching the gradient between them, enhancing the ability of semantic features to guide both SD and LLMN effectively.

\section{Supplementary Quantitative Results}

In this section, we will provide the quantitative results for each participant and include additional ablation experiments to validate the effectiveness of components in the model and the training process.

\begin{table*}[ht]
    \centering
    \renewcommand\arraystretch{1.2}
    \captionsetup{font=small}
    \small
    \caption{Results for Subject-1, 2, 5 and 7 in fMRI-to-image single-subject refining stage. }
    \scalebox{0.8}
    {
    \begin{tabular}{lccccccccc}
        \toprule
        \multicolumn{1}{c}{\multirow{2}{*}{Subject}} & \multicolumn{1}{c}{\multirow{2}{*}{Method}} & \multicolumn{4}{c}{Low-Level} & \multicolumn{4}{c}{High-Level} \\
        \cmidrule(l){3-6} \cmidrule(l){7-10}
        & & PixCorr $\uparrow$ & SSIM $\uparrow$ & AlexNet(2) $\uparrow$ & AlexNet(5) $\uparrow$ & Inception $\uparrow$ & CLIP $\uparrow$ & EffNet-B $\downarrow$ & SwAV $\downarrow$  \\
        \midrule
        \multicolumn{1}{c}{\multirow{2}{*}{Sub-1}} & Mingeye & $\mathbf{0.390}$ & ${0.337}$ & ${97.4\%}$ & ${98.7\%}$  & ${94.5\%}$ & $\mathbf{94.6\%}$ & ${0.630}$ & ${0.358}$ \\
        & \textbf{OURS} & $0.277$  & $\mathbf{0.385}$  & $\mathbf{98.8\%}$ & $\mathbf{99.3\%}$ & $\mathbf{96.2\%}$ & $94.5\%$ & $\mathbf{0.619}$  & $\mathbf{0.334}$ \\
        \midrule
        \multicolumn{1}{c}{\multirow{2}{*}{Sub-2}} & Mingeye & $\mathbf{0.318}$ & ${0.327}$ & ${95.8\%}$ & ${98.1\%}$  & ${93.2\%}$ & $\mathbf{93.7\%}$ & ${0.656}$ & ${0.368}$ \\
        & \textbf{OURS} & $0.245$  & $\mathbf{0.383}$  & $\mathbf{98.7\%}$ & $\mathbf{99.0\%}$ & $\mathbf{94.4\%}$ & $93.2\%$ & $\mathbf{0.638}$  & $\mathbf{0.348}$ \\
        \midrule
        \multicolumn{1}{c}{\multirow{2}{*}{Sub-5}} & Mingeye & $\mathbf{0.265}$ & ${0.311}$ & ${93.2\%}$ & ${97.8\%}$  & ${94.9\%}$ & $\mathbf{94.9\%}$ & ${0.628}$ & ${0.353}$ \\
        & \textbf{OURS} & $0.197$  & $\mathbf{0.366}$  & $\mathbf{94.2\%}$ & $\mathbf{98.2\%}$ & $\mathbf{94.9\%}$ & $94.3\%$ & $\mathbf{0.627}$  & $\mathbf{0.343}$ \\
        \midrule
        \multicolumn{1}{c}{\multirow{2}{*}{Sub-7}} & Mingeye & $\mathbf{0.261}$ & ${0.316}$ & ${92.3\%}$ & ${96.6\%}$  & ${92.4\%}$ & $\mathbf{93.0\%}$ & $\mathbf{0.666}$ & ${0.387}$ \\
        & \textbf{OURS} & $0.198$  & $\mathbf{0.366}$  & $\mathbf{94.5\%}$ & $\mathbf{97.0\%}$ & $\mathbf{92.5\%}$ & $91.4\%$ & $0.671$  & $\mathbf{0.375}$ \\
        \bottomrule
    \end{tabular}
    }
    \label{tab:eachsub}
\end{table*}

\subsection{Subject-Specific Results}

We present the quantitative results of the fMRI-to-image single-subject refining stage for subjects 1, 2, 5, and 7 in \cref{tab:eachsub}. The metrics for Subject-7 are slightly inferior to those of other participants, consistent with previous findings \cite{scotti2023reconstructing}. This difference could be attributed to individual variations in brain activity or subjective factors during data collection. 

Additionally, we report the quantitative results of all subjects in fMRI-to-image cross-subject pre-training stage in \cref{tab:eachsub_pretrain} and \cref{tab:eachsub_pretrain_womask}, including the the average metrics of subjects 1, 2, 5, and 7 and the average metrics of subjects 1-8.

\begin{table*}[ht]
    \centering
    \renewcommand\arraystretch{1.2}
    \captionsetup{font=small}
    \small
    \caption{Results for all subjects in the fMRI-to-image cross-subject pre-training stage, trained by replacing 5\% of the semantic features with unconditional CLIP embeddings corresponding to empty tokens. }
    \scalebox{0.8}
    {
    \begin{tabular}{lcccccccc}
        \toprule
        \multicolumn{1}{c}{\multirow{2}{*}{Subject}} & \multicolumn{4}{c}{Low-Level} & \multicolumn{4}{c}{High-Level} \\
        \cmidrule(l){2-5} \cmidrule(l){6-9}
        & PixCorr $\uparrow$ & SSIM $\uparrow$ & AlexNet(2) $\uparrow$ & AlexNet(5) $\uparrow$ & Inception $\uparrow$ & CLIP $\uparrow$ & EffNet-B $\downarrow$ & SwAV $\downarrow$  \\
        \midrule
        Sub-1 & 0.167 & 0.350 & 95.1\% & 97.7\% & 89.1\% & 90.7\% & 0.698 & 0.399 \\
        Sub-2 & 0.154 & 0.350 & 94.1\% & 96.8\% & 88.8\% & 88.2\% & 0.721 & 0.412 \\
        Sub-3 & 0.115 & 0.342 & 88.3\% & 93.7\% & 85.6\% & 86.2\% & 0.760 & 0.447 \\
        Sub-4 & 0.105 & 0.346 & 86.3\% & 92.8\% & 84.5\% & 86.1\% & 0.758 & 0.448 \\
        Sub-5 & 0.124 & 0.346 & 87.8\% & 95.2\% & 89.9\% & 90.5\% & 0.713 & 0.414 \\
        Sub-6 & 0.108 & 0.343 & 86.1\% & 92.4\% & 86.0\% & 88.0\% & 0.751 & 0.439 \\
        Sub-7 & 0.117 & 0.350 & 88.6\% & 93.1\% & 85.4\% & 86.2\% & 0.757 & 0.443 \\
        Sub-8 & 0.096 & 0.345 & 84.6\% & 90.4\% & 80.3\% & 82.1\% & 0.798 & 0.479 \\
        \midrule
        Avg (Sub 1,2,5 \& 7) & 0.141 & 0.349 & 91.4\% & 95.7\% & 88.3\% & 88.9\% & 0.722 & 0.417 \\
        Avg (Sub 1-8) & 0.123 & 0.347 & 88.9\% & 94.0\% & 86.2\% & 87.3\% & 0.744 & 0.435 \\
        \bottomrule
    \end{tabular}
    }
    \label{tab:eachsub_pretrain}
\end{table*}

\begin{table*}[ht]
    \centering
    \renewcommand\arraystretch{1.2}
    \captionsetup{font=small}
    \small
    \caption{Results for all subjects in the fMRI-to-image cross-subject pre-training stage without using unconditional CLIP embeddings to replace the semantic features. }
    \scalebox{0.8}
    {
    \begin{tabular}{lcccccccc}
        \toprule
        \multicolumn{1}{c}{\multirow{2}{*}{Subject}} & \multicolumn{4}{c}{Low-Level} & \multicolumn{4}{c}{High-Level} \\
        \cmidrule(l){2-5} \cmidrule(l){6-9}
        & PixCorr $\uparrow$ & SSIM $\uparrow$ & AlexNet(2) $\uparrow$ & AlexNet(5) $\uparrow$ & Inception $\uparrow$ & CLIP $\uparrow$ & EffNet-B $\downarrow$ & SwAV $\downarrow$  \\
        \midrule
        Sub-1 & 0.168 & 0.346 & 94.5\% & 96.8\% & 87.7\% & 89.0\% & 0.726 & 0.419 \\
        Sub-2 & 0.138 & 0.340 & 93.9\% & 96.4\% & 85.8\% & 87.0\% & 0.752 & 0.438 \\
        Sub-3 & 0.103 & 0.327 & 86.6\% & 92.1\% & 83.7\% & 84.9\% & 0.776 & 0.461 \\
        Sub-4 & 0.088 & 0.325 & 85.0\% & 91.2\% & 81.8\% & 84.2\% & 0.782 & 0.470 \\
        Sub-5 & 0.111 & 0.329 & 87.0\% & 94.2\% & 87.3\% & 88.4\% & 0.735 & 0.431 \\
        Sub-6 & 0.094 & 0.322 & 84.9\% & 90.8\% & 84.2\% & 85.3\% & 0.768 & 0.458 \\
        Sub-7 & 0.110 & 0.331 & 85.8\% & 91.9\% & 82.8\% & 84.3\% & 0.771 & 0.457 \\
        Sub-8 & 0.087 & 0.326 & 83.0\% & 88.8\% & 77.2\% & 79.7\% & 0.818 & 0.494 \\
        \midrule
        Avg (Sub 1,2,5 \& 7) & 0.132 & 0.337 & 90.3\% & 94.8\% & 85.9\% & 87.2\% & 0.746 & 0.436 \\
        Avg (Sub 1-8) & 0.112 & 0.331 & 87.6\% & 92.8\% & 83.8\% & 85.4\% & 0.766 & 0.453 \\
        \bottomrule
    \end{tabular}
    }
    \label{tab:eachsub_pretrain_womask}
\end{table*}

\subsection{Ablation Study}

To assess the efficacy of both the hyperparameter $\lambda$ and the different variants of the Low-Level Manipulation Network in our model, we shows the findings from our ablation studies in \cref{tab:ablation_lambda} and \cref{tab:ablation_llmn}.

\noindent
{\bf Impact of hyperparameter $\lambda$.} 
We vary $\lambda$ to assess its impact on the results. For efficiency, we train for 30,000 iterations during single-person fine-tuning without full convergence. \cref{tab:ablation_lambda} shows that $\lambda=0.1$ yields the best results across nearly all metrics. Values of $\lambda$ within 0.05 to 0.15 demonstrate robustness. However, excessively high values of $\lambda$ (0.5) notably degrade the low-level performance of the generated results, as an overly weighted semantic loss compromises the low-level manipulation network's performance.

\begin{table*}[ht]
    \centering
    \captionsetup{font=small}
    \renewcommand\arraystretch{1.0}
    \setlength{\tabcolsep}{1pt}
    \small
     \caption{ \small 
    Ablation study on the impact of hyperparameter $\lambda$ (30,000 iterations). 
    }
    \scalebox{0.83}
    {\begin{tabular}{lcccccccccc}
        \toprule
        \multicolumn{2}{c}{Method}
        & \multicolumn{4}{c}{Low-Level} & \multicolumn{4}{c}{High-Level} \\
        \cmidrule(l){1-2} \cmidrule(l){3-6} \cmidrule(l){7-10}
        Model & $\lambda$ & PixCorr $\uparrow$ & SSIM $\uparrow$ & AlexNet(2) $\uparrow$ & AlexNet(5) $\uparrow$ & Inception $\uparrow$ & CLIP $\uparrow$ & EffNet-B $\downarrow$ & SwAV $\downarrow$  \\
        \midrule
        NeuroPictor & 0.05  & 0.230  & 0.375  & \textbf{98.7\%} & 99.1\% & \textbf{95.5\%} & 94.3\% & 0.634 & 0.348 \\
        NeuroPictor & \textbf{0.10} & \textbf{0.238}  & \textbf{0.383}  & 98.5\% & \textbf{99.3\%} & \textbf{95.5\%} & \textbf{94.5\%} & \textbf{0.624} & \textbf{0.342} \\
        NeuroPictor & 0.15  & 0.227  & 0.373  & 98.3\% & 99.1\% & 95.1\% & 94.4\% & 0.625 & 0.344\\
        NeuroPictor & 0.25  & 0.219  & 0.369  & 98.2\% & 99.1\% & 94.9\% & 93.6\% & 0.631 & 0.348\\
        NeuroPictor & 0.50  & 0.204  & 0.369  & 97.6\% & 98.9\% & 94.4\% & 93.8\% & 0.634 & 0.352\\
        \bottomrule
    \end{tabular}}
    \label{tab:ablation_lambda}
\end{table*}

\begin{table*}[ht]
    \centering
    \captionsetup{font=small}
    \renewcommand\arraystretch{1.0}
    \setlength{\tabcolsep}{1pt}
    \small
     \caption{ \small 
    Ablation study on the variants of the Low-Level Manipulation Network. All results are obtained without fMRI-to-image cross-subject pre-training; instead, we directly conduct fMRI-to-image single-subject refining.
    }
    \scalebox{0.76}
    {\begin{tabular}{lcccccccccc}
        \toprule
        \multicolumn{2}{c}{Method}
        & \multicolumn{4}{c}{Low-Level} & \multicolumn{4}{c}{High-Level} \\
        \cmidrule(l){1-2} \cmidrule(l){3-6} \cmidrule(l){7-10}
        Model & LLMN & PixCorr $\uparrow$ & SSIM $\uparrow$ & AlexNet(2) $\uparrow$ & AlexNet(5) $\uparrow$ & Inception $\uparrow$ & CLIP $\uparrow$ & EffNet-B $\downarrow$ & SwAV $\downarrow$  \\
        \midrule
        NeuroPictor & w/o init & 0.024  & 0.324  & 50.8\% & 50.4\% & 49.3\% & 51.2\% & 0.979  & 0.638 \\
        NeuroPictor & w/o cross att & 0.179  & \textbf{0.398}  & 96.2\% & 97.0\% & 91.3\% & 90.7\% & 0.722  & 0.426 \\
        NeuroPictor & full & \textbf{0.265} & 0.368 & \textbf{98.2\%} & \textbf{99.1\%} & \textbf{94.7\%} & \textbf{93.4\%} & \textbf{0.641} & \textbf{0.342} \\
        \bottomrule
    \end{tabular}}
    \label{tab:ablation_llmn}
\end{table*}

\noindent
{\bf Low-Level Manipulation Network.} 
As shown in \cref{tab:ablation_llmn}, we evaluate the effectiveness of the Low-Level Manipulation Network from two aspects. On the one hand, we initialize the UNet in the LLMN without using weights from Stable Diffusion (SD), resulting in significantly poorer trained results. On the other hand, we remove the cross-attention layer from the UNet in the LLMN (i.e., semantic features no longer guide the LLMN), and simultaneously remove the noise latent $z_T$, resulting in a variant of the LLMN. The results indicate a significant decline in performance in terms of high-level features for this variant.

\subsection{Influence of Dataset prior}
\begin{figure*}[ht]
    \centering
    \tiny\includegraphics[width=1\linewidth]{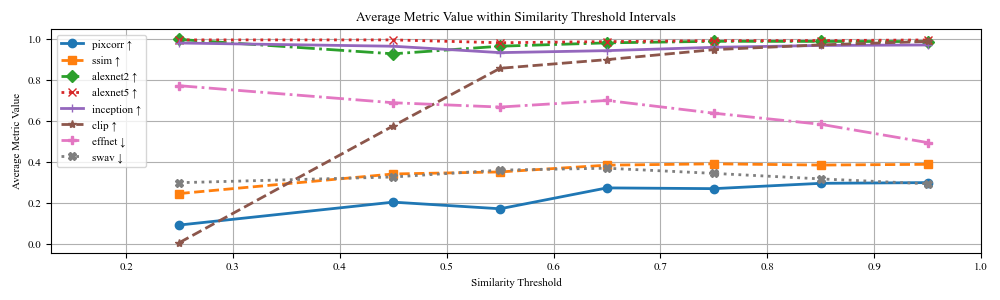}
    \caption{Averaged Metric Value within Similarity Threshold Intervals. The x-axis represents the similarity between test images and training images. The y-axis indicates the performance of the fMRI reconstruction results corresponding to these test images.
    \label{fig:sim} }
\end{figure*}

To quantify the influence of data prior, we visualize the similarity (maximum similarity of DINO v2 features) of the testing image to the training set and its reconstruction performance in \cref{fig:sim}.
Many of the metrics are robust against the similarity, especially for the similarity is larger than 50\%. Indeed, more similar testing examples enjoy better performance. By correlation coefficients, data similarity
affects high-level features (0.24) more than low-level features (0.09).

\subsection{Training corpus retrieval V.S. fMRI reconstruction.}
We retrieve the most similar images from the training dataset, and subsequently compare our actual reconstruction from the fMRI signals with the retrieval of images from the training corpus. 
DINO-V2 and fMRI features are used to retrieval from the training corpus. For each test image, we calculate its similarity with the training corpus based on either its DINO-V2 features or the encoded fMRI features. The most similar training image is chosen as the retrieval outcome. The results in \cref{tab:retirval} show that retrieved results performs better on high-level metrics but worse in low-level metrics.
This verifies that our method performs fMRI reconstruction as low-level information are unseen during training.

\begin{table*}[ht]
    \centering
    \captionsetup{font=small}
    \renewcommand\arraystretch{1.0}
    \setlength{\tabcolsep}{1pt}
    \small
     \caption{ \small 
    Comparison between training corpus retrieval and fMRI reconstruction
    }
    \scalebox{0.76}
    {\begin{tabular}{lccccccccc}
        \toprule
        Method & Retrieval Set & PixCorr $\uparrow$ & SSIM $\uparrow$ & AlexNet(2) $\uparrow$ & AlexNet(5) $\uparrow$ & Inception $\uparrow$ & CLIP $\uparrow$ & EffNet-B $\downarrow$ & SwAV $\downarrow$  \\
        \midrule
        \multirow{2}{*}{DINO V2} & Sub-1 (9k)   & 0.133 & 0.311 & 0.782 & 0.914 & 0.982 & 0.983 & 0.544 & 0.342 \\ 
        & Sub-all (67k) & 0.167 & 0.322 & 0.834 & 0.941 & 0.991 & 0.991 & 0.488 & 0.302 \\
        \midrule
        FMRI Feature & Sub-1 (9k)    & 0.121 & 0.323 & 0.785 & 0.885 & 0.881 & 0.896 & 0.717 & 0.437 \\
        \midrule
        Reconstruction & - & 0.277 & 0.385 & 0.988 & 0.993 & 0.962 & 0.945 & 0.619 & 0.334 \\
        \bottomrule
    \end{tabular}}
    \label{tab:retirval}
\end{table*}

\subsection{Perturbing cortex across individuals.} 
We replaced fMRI signals from specific regions of subject-1 with those from other subjects. 
We chose three sets of regions, as in the \cref{tab:location}, with each set having a similar number of pixels in the 2-D map.  
Replacing higher-level visual cortex regions had a greater impact. This may suggests that lower-level visual cortex regions are more consistent across individuals, while higher-level regions are more unique.

\begin{table}[ht] \small
    \centering
    \setlength{\tabcolsep}{1pt}
    \renewcommand\arraystretch{1.0}
    \caption{
    Perturbing cortex across individuals by replacing fMRI signals from specific regions of subject-1 with those from subject-2.
    }
    \scalebox{0.76}{
    \begin{tabular}{lccccccccc}
        \toprule
        Region & RoIs & PixCorr $\uparrow$ & SSIM $\uparrow$ & AlexNet(2) $\uparrow$ & AlexNet(5) $\uparrow$ & Inception $\uparrow$ & CLIP $\uparrow$ & EffNet-B $\downarrow$ & SwAV $\downarrow$  \\
        \midrule
        Primary Visual     & V1    & 0.176 & 0.357 & 0.962 & 0.985 & 0.944 & 0.930 & 0.647 & 0.358 \\
        Early Visual       & V2-V4 & 0.223 & 0.371 & 0.971 & 0.983 & 0.921 & 0.918 & 0.683 & 0.380 \\
        Some Higher Visual & Others & 0.219 & 0.371 & 0.968 & 0.972 & 0.845 & 0.825 & 0.794 & 0.451 \\
        \midrule
        None & - & 0.277 & 0.385 & 0.988 & 0.993 & 0.962 & 0.945 & 0.619 & 0.334 \\
        \bottomrule
    \end{tabular}
    }
    \label{tab:location}
\end{table}

\section{Supplementary Visualizations}

In this section, we will present additional visual results for each subject and use visual examples to illustrate the impact of the components and training strategies in our method.

\subsection{Visualization of mismatched high-level and low-level features}

\begin{figure*}[htbp]
\begin{centering}
\includegraphics[width=0.82\linewidth]{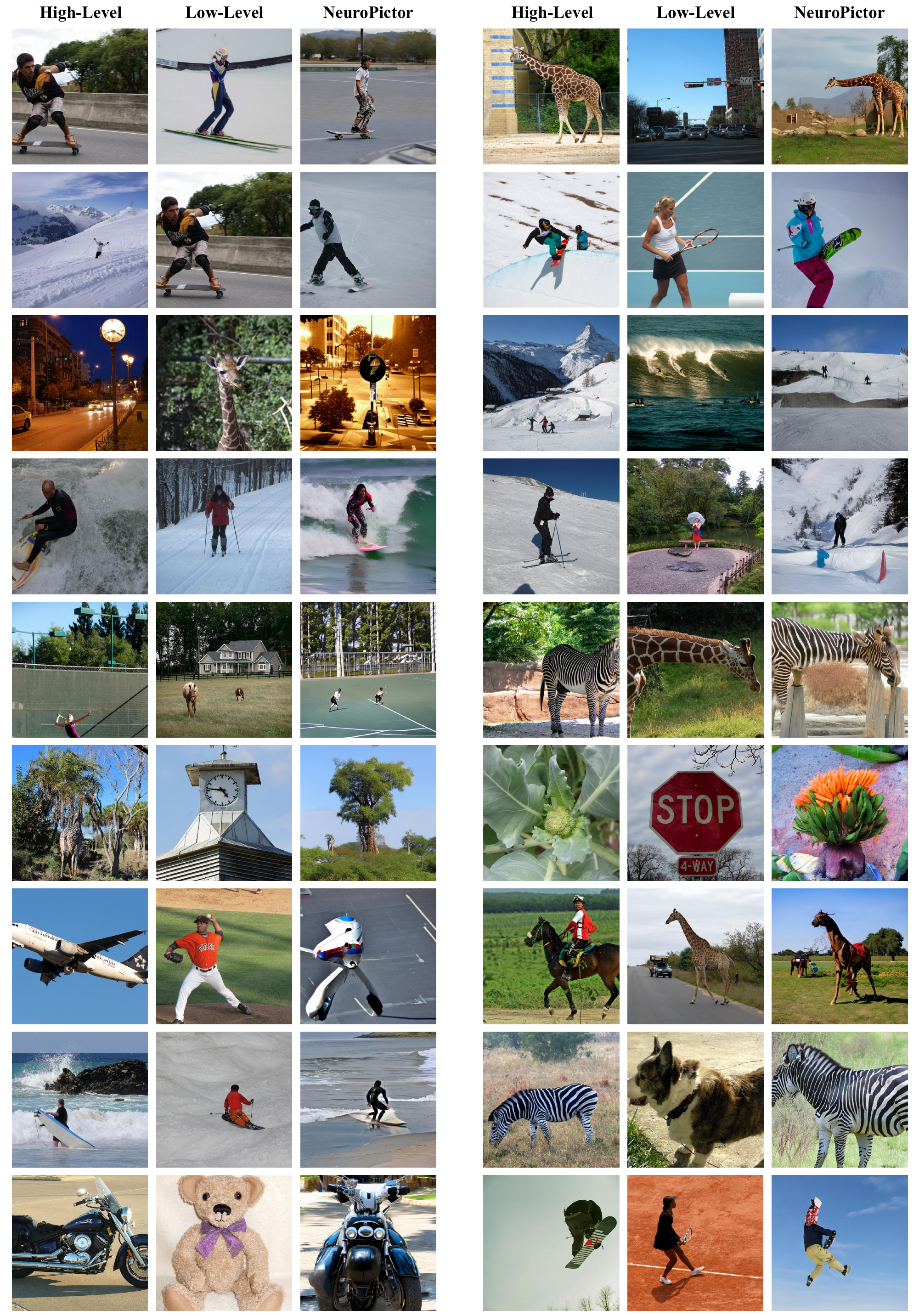}
\par\end{centering}
 \caption{Visualization of mismatched high-level and low-level features. \label{fig:supp_exchange}}
\end{figure*}

We utilize two distinct fMRI sources, with one providing high-level features and the other providing low-level features. 
As illustrated in \cref{fig:supp_exchange}, this approach results in the generation results where the visual structures, object positions, and human poses align with the low-level source visual stimuli, while maintaining semantic consistency with the high-level source.
This disentanglement of high-level meaning and low-level structure in fMRI signals indeed opens up intriguing possibilities, suggesting the potential for fMRI-driven image editing.

\subsection{Subject-Specific Visualizations}

\begin{figure*}[htbp]
\begin{centering}
\includegraphics[width=0.97\linewidth]{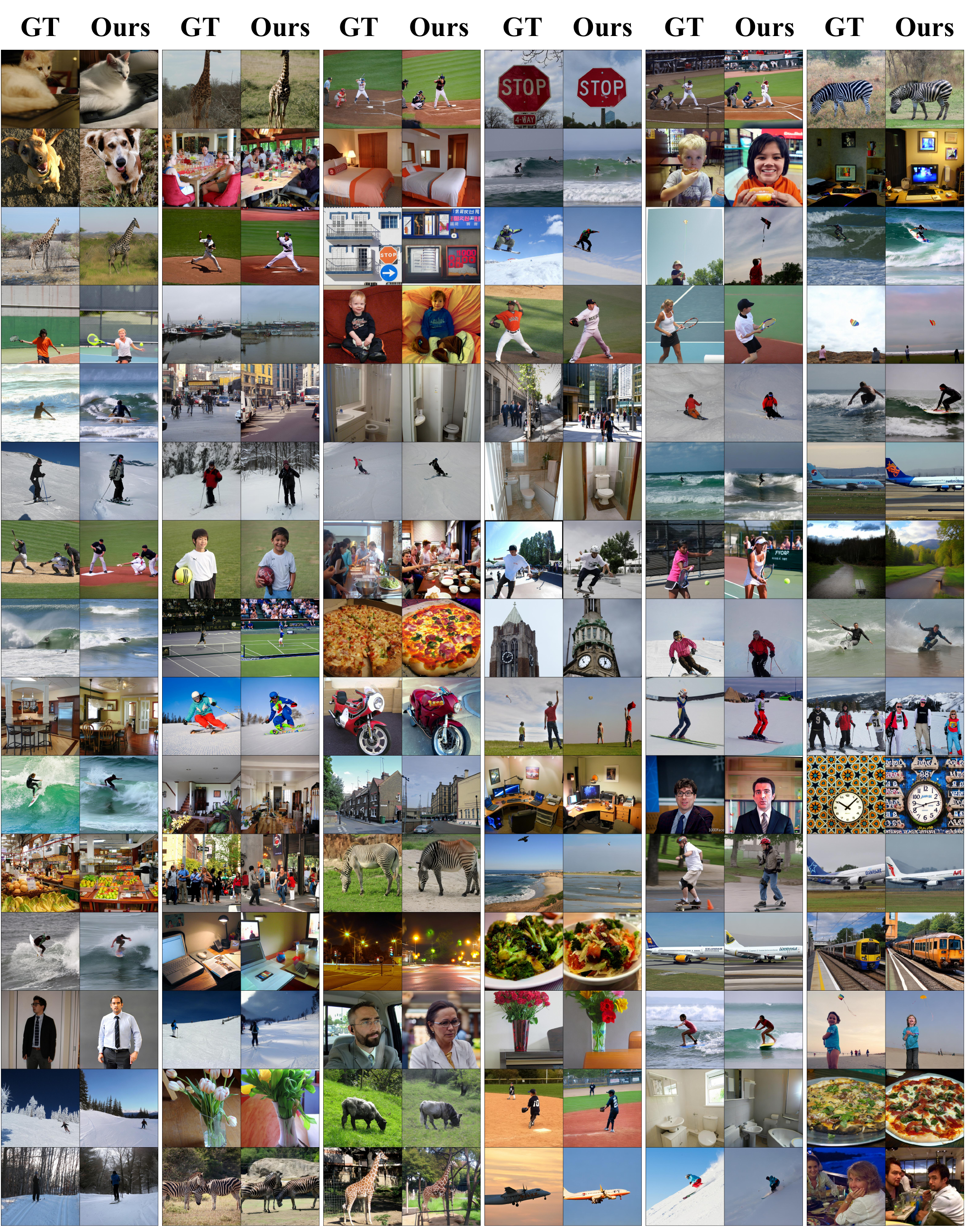}
\par\end{centering}
 \caption{Visualizations of fMRI-to-image reconstruction for Subject-1. \label{fig:sub1}}
\end{figure*}

\begin{figure*}[htbp]
\begin{centering}
\includegraphics[width=0.97\linewidth]{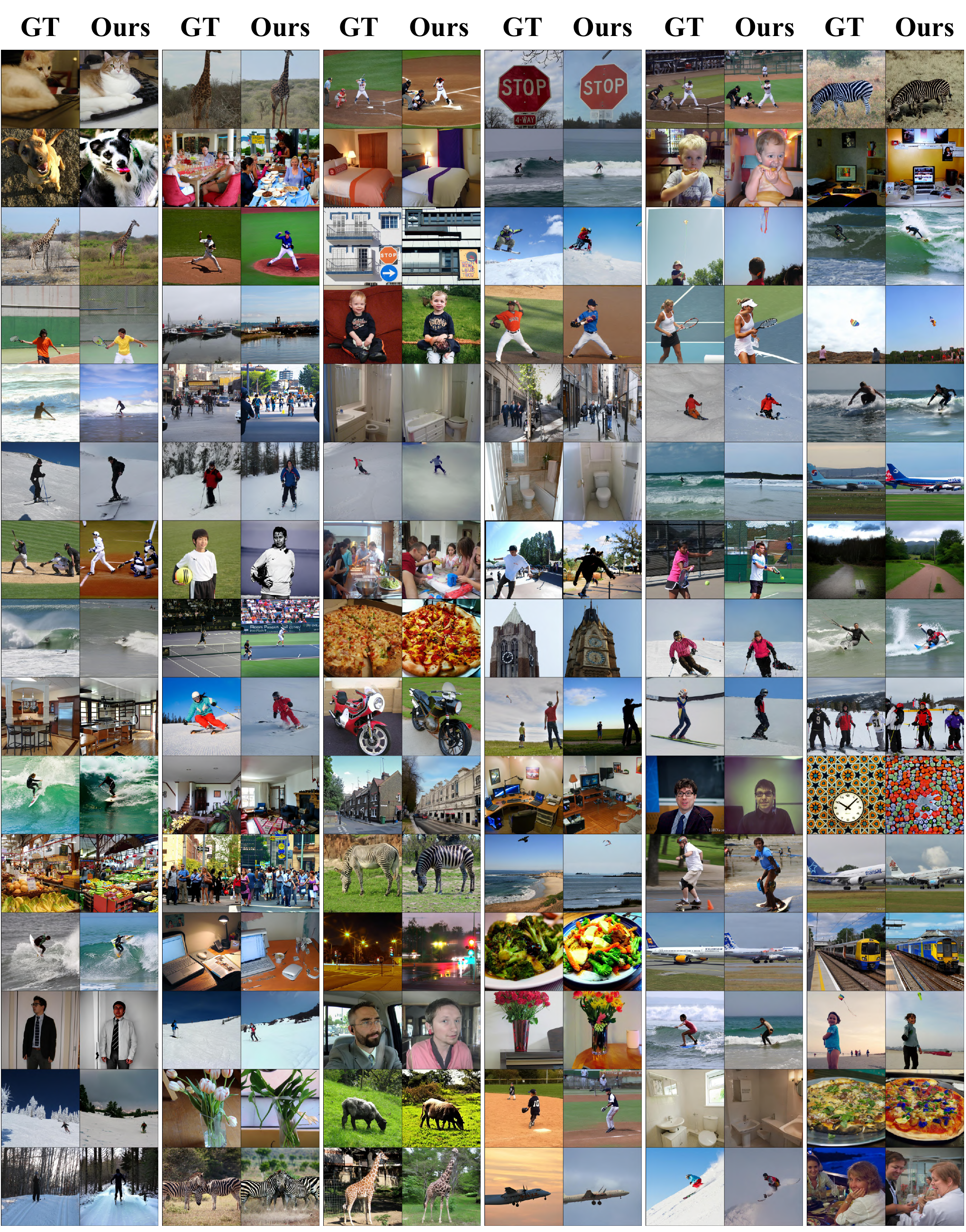}
\par\end{centering}
 \caption{Visualizations of fMRI-to-image reconstruction for Subject-2. \label{fig:sub2}}
\end{figure*}

\begin{figure*}[htbp]
\begin{centering}
\includegraphics[width=0.97\linewidth]{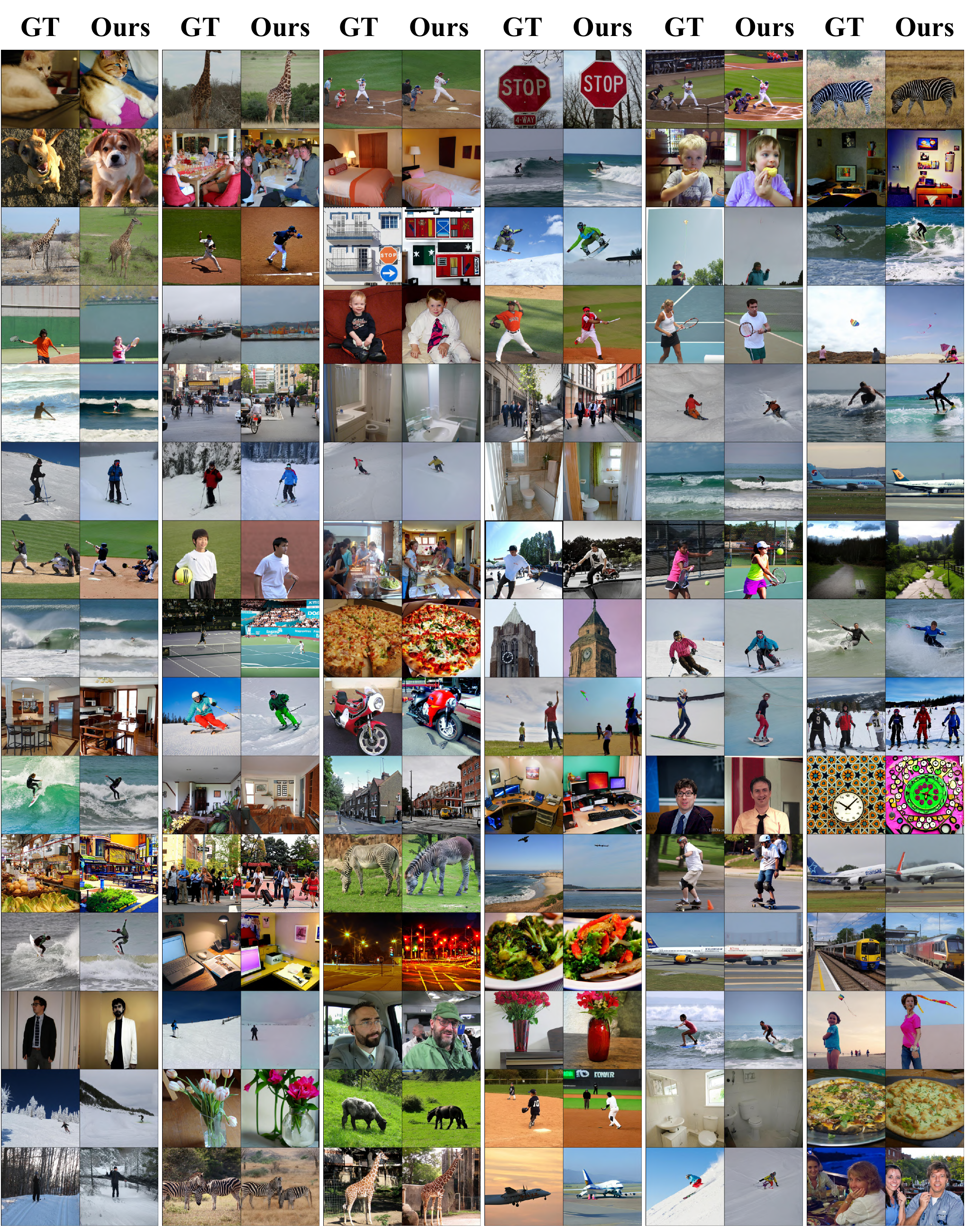}
\par\end{centering}
 \caption{Visualizations of fMRI-to-image reconstruction for Subject-5. \label{fig:sub5}}
\end{figure*}

\begin{figure*}[htbp]
\begin{centering}
\includegraphics[width=0.97\linewidth]{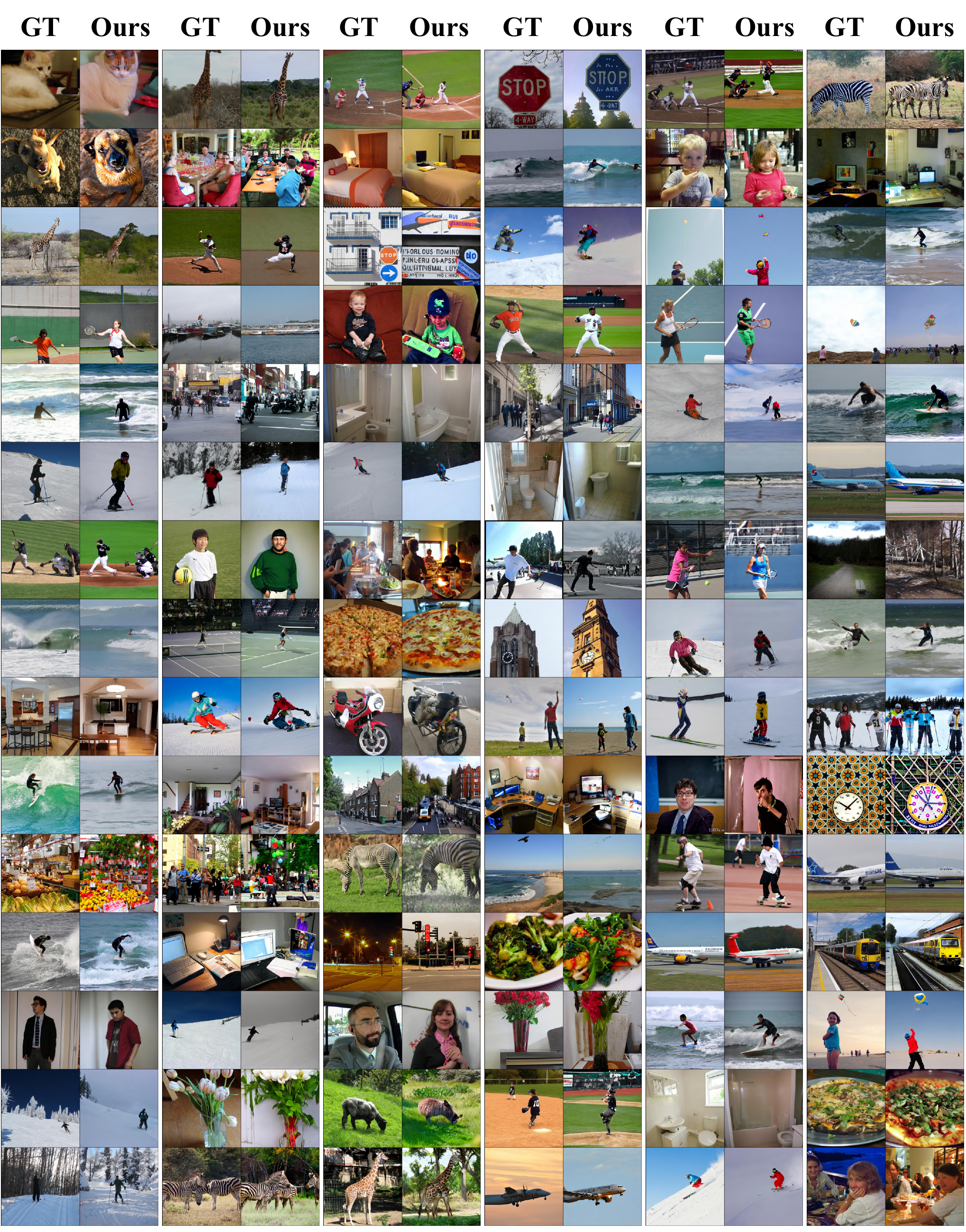}
\par\end{centering}
 \caption{Visualizations of fMRI-to-image reconstruction for Subject-7. \label{fig:sub7}}
\end{figure*}

We provide additional visual results for fMRI-to-image reconstruction, including visualizations for the four participants (Subjects 1, 2, 5, and 7) used for evaluation. The results for these four subjects (Subjects 1, 2, 5, and 7) are visualized separately in \cref{fig:sub1}, \cref{fig:sub2}, \cref{fig:sub5} and \cref{fig:sub7}. All results are selected from the test set, with each participant having the same ground truth images.

\subsection{Visualizations for Ablation Study}

\begin{figure*}[htbp]
\begin{centering}
\includegraphics[width=1.0\linewidth]{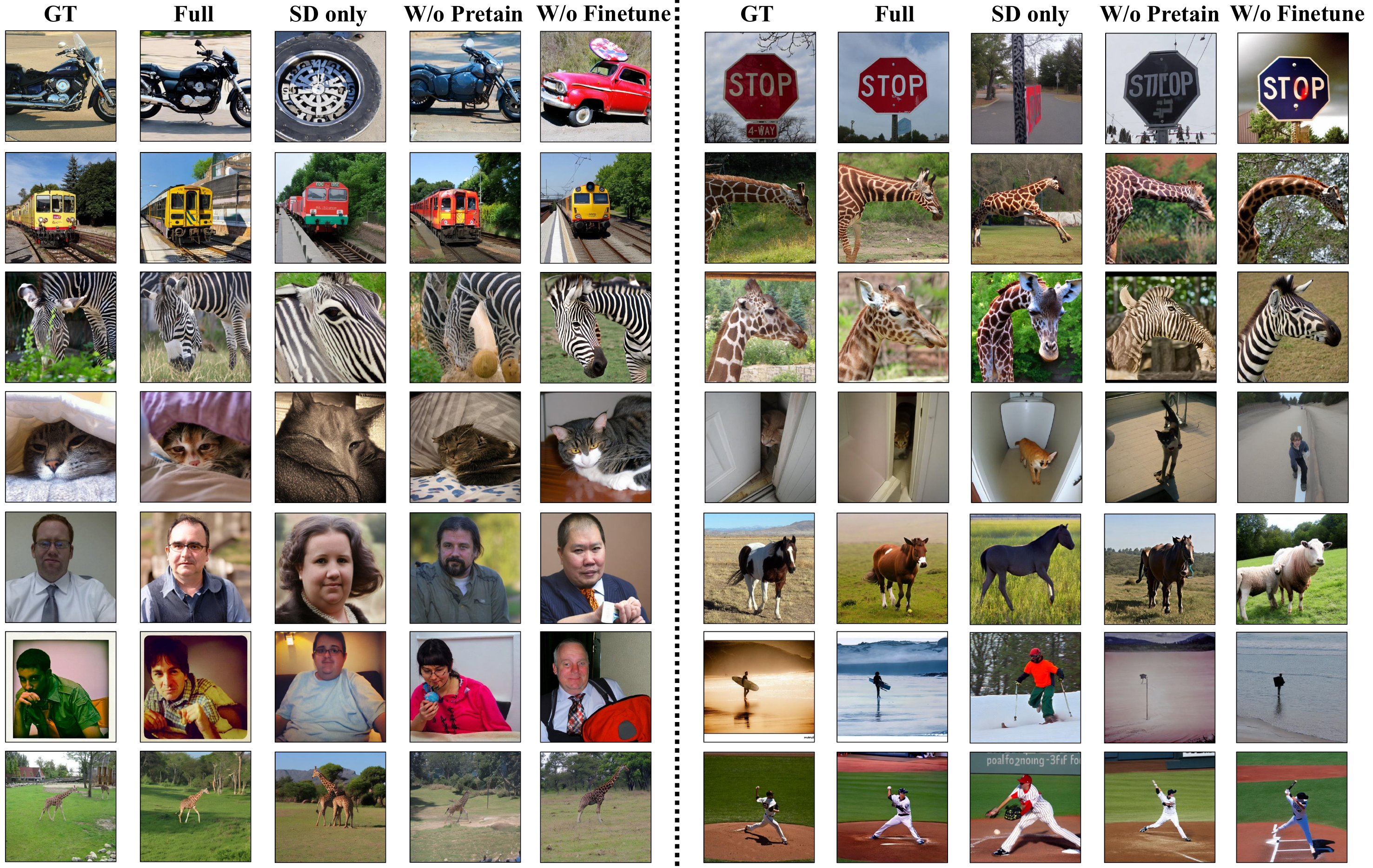}
\par\end{centering}
 \caption{Visualizations for ablation study. `` SD only" signifies the results obtained by using only Stable Diffusion without adding Low-Level Manipulation Network. `` Full" denotes the complete model. `` W/o Pretrain" indicates training the fMRI-to-image model directly on Subject-1 without pretraining. `` W/o Finetune" represents the results of training the fMRI-to-image model on all participants without further finetuning on Subject-1. \label{fig:supp_ab}}
\end{figure*}

We provide visual examples to validate the effectiveness of the Low-Level Manipulation Network branch, pretraining strategy, and finetuning strategy. The visualizations are depicted in \cref{fig:supp_ab}. "SD only" denotes results obtained using only Stable Diffusion without incorporating the Low-Level Manipulation Network. "Full" represents the complete model. "W/o Pretrain" indicates training the fMRI-to-image model directly on Subject-1 without pretraining. "W/o Finetune" denotes results of training the fMRI-to-image model on all participants without additional finetuning on Subject-1. As observed, compared to the full model, using only Stable Diffusion results in reconstructions that roughly match in terms of semantics. The quality of images generated solely through finetuning is poor. When only pretraining is performed, there are instances of semantic errors, and the reconstructed images may not sufficiently match the original images in low-level details.

\subsection{Failure Cases}

Examples of genetation failures of our NeuroPictor are shown in \cref{fig:supp_fail}. Despite the advancements, our method still exhibits limitations, occasionally confusing semantically similar concepts such as cats and dogs, or males and females. Additionally, for some complex scenes, our generated results may occasionally omit certain elements, such as generating only dogs without including cars or mirrors, or generating polar bears without incorporating the presence of water surfaces.

\section{Computational Cost}

We compare the parameters, inference time, and memory usage of our NeuroPictor with those of Mindeye \cite{scotti2023reconstructing}. 
We perform inference using an NVIDIA RTX A6000 GPU, where the inference time denotes the average time taken to generate a single image. For fair comparison, both models are configured with float32 precision and execute 50 DDIM \cite{song2020denoising} sampling timesteps. In the original Mindeye approach, 16 samples are generated for each fMRI input, from which the best one is selected. We omit this step when measuring inference time for fairness.
As shown in \cref{tab:runtime}, we achieve state-of-the-art results with lower inference-stage computational costs compared to Mindeye. 

\begin{table*}[h]
    \centering
    \captionsetup{font=small}
    \small
     \caption{ \small 
    Comparison of the computation cost. 
    }
    \begin{tabular}{cccccccc}
         \toprule
         \multicolumn{1}{c}{Model} & \multicolumn{1}{c}{Params (B)} & \multicolumn{1}{c}{Memory (GB)} & \multicolumn{1}{c}{Infer Time (s)} \\
         \midrule
         Mindeye & 2.9 & 33.4 & 10.19 \\
         Ours & 2.0 & 13.6 & 7.26  \\
         \bottomrule
    \end{tabular}
    \label{tab:runtime}
\end{table*}

\section{Limitations and Future work}

\begin{figure*}[htbp]
\begin{centering}
\includegraphics[width=1.0\linewidth]{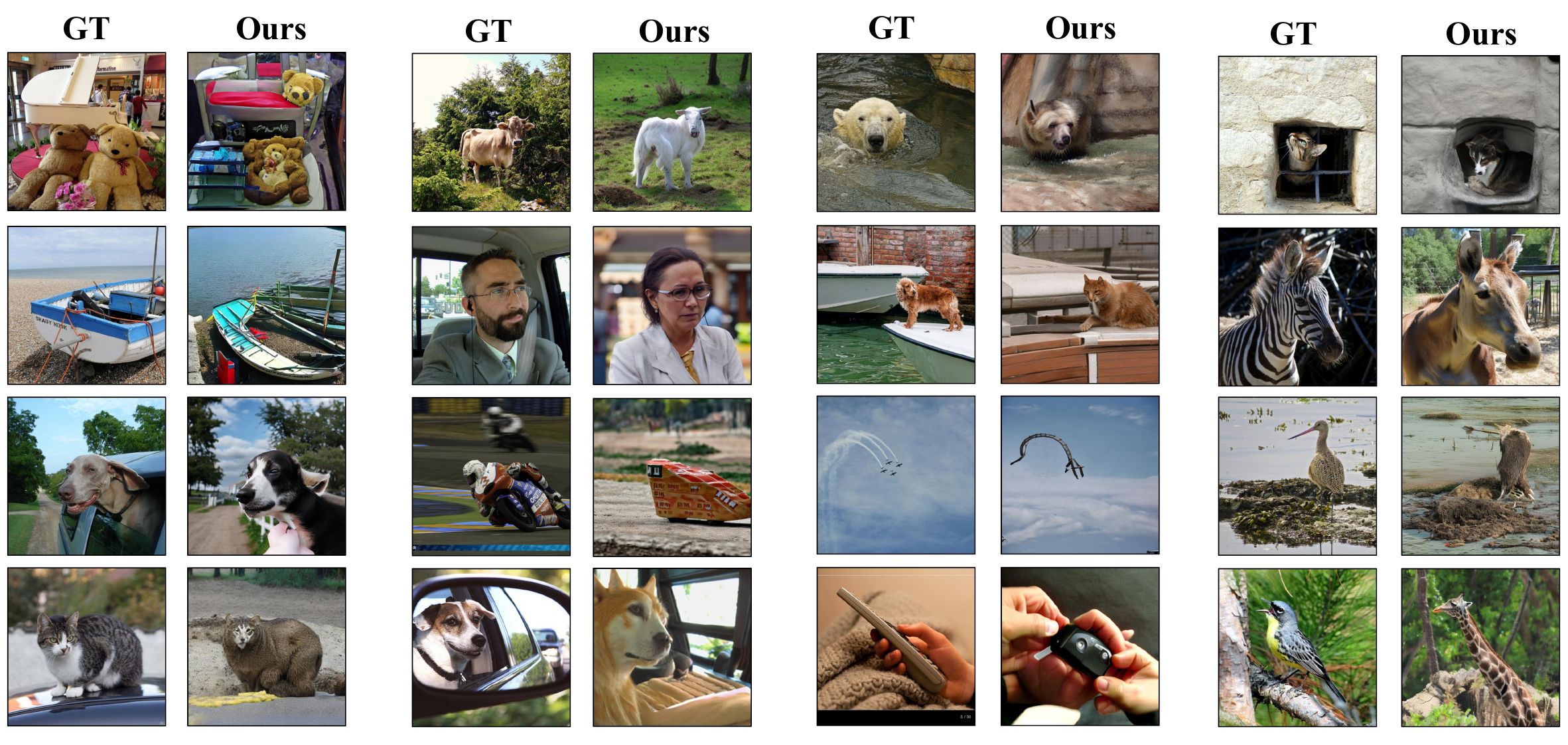}
\par\end{centering}
 \caption{Visualizations for failure cases. \label{fig:supp_fail}}
\end{figure*}

Although our NeuroPictor has shown improvements in fMRI-to-image decoding for individual subjects by leveraging pretraining on multiple-individual fMRI-image pairs, there are still several limitations:
(1) The single-subject fine-tuning strategy we adopted requires fine-tuning the entire model, which incurs substantial computational costs. Therefore, it is worthwhile to explore more efficient strategies for single-subject fine-tuning. This could involve learning additional \textit{special embeddings} for different individuals or fine-tuning only lightweight modules, rather than the entire model.
(2) Currently, there are some challenges in generalizing directly to unseen subjects (\ie, cross-subject) or new experimental setups (e.g., cross-device). In this regard, future work could focus on developing methods to rapidly fine-tune a multi-individual pretrained model for unseen individuals.
(3) The high-level and low-level branches of NeuroPictor are trained in a coupled manner. Although the SD model priors and the fMRI-to-text branch help to decouple semantic information and low-level details, this is not optimal. There are some semantic errors as shown in \cref{fig:supp_fail}. New strategies could be considered to better decode multi-level information.
(4) We only use specific visual cortices for decoding. Incorporating other brain regions is also worth considering, as it may provide new insights.

\end{document}